\documentclass[journal]{IEEEtran}
\IEEEoverridecommandlockouts
\usepackage{cite}
\usepackage{amsmath,amssymb,amsfonts}
\usepackage{algorithmic}
\usepackage[caption=false,font=footnotesize,labelfont=rm,textfont=rm]{subfig}
\usepackage{graphicx}
\usepackage{textcomp}
\usepackage{xcolor}
\usepackage{float}
\usepackage{url}

\newcommand{\sh}[1]{{\color{black}#1}}

\begin{document}
\markboth{Journal of \LaTeX\ Class Files,~Vol.~18, No.~9, September~2020}%
{Exploring Spatial-Temporal Representation via Star Graph for mmWave Radar-based Human Activity Recognition}
 
\title{Exploring Spatial-Temporal Representation via Star Graph for mmWave Radar-based Human Activity Recognition}

\author{
\IEEEauthorblockN{
Senhao~Gao,~\IEEEmembership{Student Member,~IEEE}, 
Junqing~Zhang,~\IEEEmembership{Senior Member,~IEEE}, 
Luoyu~Mei,
Shuai~Wang,~\IEEEmembership{Senior~Member,~IEEE}, 
and
Xuyu~Wang,~\IEEEmembership{Member,~IEEE}
}

\thanks{Manuscript received xxx; revised xxx; accepted xxx. Date of publication xxx; date of current version xxx. The work was partly supported by the UK Engineering and Physical Sciences Research Council (EPSRC) New Investigator Award under grant ID EP/V027697/1. The review of this paper was coordinated by xxx. 
(\textit{Corresponding author: Junqing Zhang})}
\IEEEcompsocitemizethanks{
\IEEEcompsocthanksitem S.~Gao and J.~Zhang are with the School of Computer Science and Informatics, University of Liverpool, Liverpool, L69 3DR, United Kingdom. (email: \{Senhao.Gao, Junqing.Zhang\}@liverpool.ac.uk)
}
\IEEEcompsocitemizethanks{L.~Mei is with the School of Computer Science and Engineering, Southeast University, Nanjing, Jiangsu, 210096, China. He is also with City University of Hong Kong, Hong Kong. (email: lymei-@seu.edu.cn)}
\IEEEcompsocitemizethanks{S.~Wang is with the School of Computer Science and Engineering, Southeast University, Nanjing, Jiangsu, 210096, China. (email: shuaiwang@seu.edu.cn)}
\IEEEcompsocitemizethanks{X.~Wang is with the Knight Foundation School of Computing and Information Sciences, Florida International University, Miami, FL 33199,  USA. (email: xuywang@fiu.edu)}
\thanks{Color versions of one or more of the figures in this paper are available online at http://ieeexplore.ieee.org.}
\thanks{Digital Object Identifier xxx}
}

\maketitle
\begin{abstract}
Human activity recognition (HAR) requires extracting accurate spatial-temporal features with human movements. A mmWave radar point cloud-based HAR system suffers from sparsity and variable-size problems due to the physical features of the mmWave signal. 
Existing works usually borrow the preprocessing algorithms for the vision-based systems with dense point clouds, which may not be optimal for mmWave radar systems. 
In this work, we proposed a graph representation with a discrete dynamic graph neural network (DDGNN) to explore the spatial-temporal representation of human movement-related features. Specifically, we designed a star graph to describe the high-dimensional relative relationship between a manually added static center point and the dynamic mmWave radar points in the same and consecutive frames. 
We then adopted DDGNN to learn the features residing in the star graph with variable sizes.
Experimental results demonstrated that our approach outperformed other baseline methods using real-world HAR datasets. Our system achieved an overall classification accuracy of 94.27\%, which gets the near-optimal performance with a vision-based skeleton data accuracy of 97.25\%. We also conducted an inference test on Raspberry Pi~4 to demonstrate its effectiveness on resource-constraint platforms. \sh{ We provided a comprehensive ablation study for variable DDGNN structures to validate our model design. Our system also outperformed three recent radar-specific methods without requiring resampling or frame aggregators. } 
\end{abstract}


\section{Introduction} 
Human activity recognition (HAR) is an important research topic, which has enabled transformative applications, such as human-computer interaction systems, robot control, etc~\cite{liu2019wireless}. 
Currently, HAR systems are mainly based on vision sensors, e.g., RGB or depth cameras~\cite{chen2021deep}. However, camera-based HAR systems have privacy concerns, significantly limiting their applications in privacy-sensitive environments, such as bathrooms. Recently, millimeter wave (mmWave) radar \cite{shastri2022review} has become a good complementary technology because it does not violate privacy.
Hence, mmWave radar has attracted increasing research interest and effort~\cite{singh2019radhar, yu2022noninvasive, chen203universal, cui2024milipoint, zhang2023survey}.

\subsection{Vision Sensor-based and mmWave Radar-based HAR}
Both vision sensors and mmWave radar can provide geometric and distance information by capturing reflected signals and measuring the ambient environment. 
Both can output point clouds, which are a set of points containing the 3D location of the detected objects. 

 Currently, popular vision sensors include Microsoft Kinect \cite{zhang2012microsoft} and LIDAR \cite{alam2021palmar}, etc. Kinect uses infrared lights with an RGB camera, and LIDAR uses laser pulses. They both can scan the ambient environment at a very high frequency and thus can provide a dense point cloud set that represents the spatial characteristics of the environment \cite{chen2021deep}, which normally has over 10,000 points in one frame. 

Instead, mmWave radar uses reflected mmWave band (30~GHz-300~GHz) signals to detect objects in the environment. Unlike the dense point cloud of vision sensors, due to the physical features of the mmWave band signal, mmWave radar suffers from multipath interference and is sensitive to high-reflective surfaces, e.g., metals~\cite{zhang2023survey}. After eliminating these environmental noises and interference signals (or ghost signals), the mmWave radar point cloud is sparse, and the size is variable in different time periods, which normally have around 10-50 points in one frame~\cite{venon2022millimeter}. 

Deep learning has been widely used in both vision-based and mmWave radar-based HAR~\cite{gu2021survey} thanks to its excellent classification capability. Deep learning-based HAR usually consists of preprocessing algorithms and deep learning engines. The preprocessing is designed to enhance the human-related information in the raw data, while the deep learning engines extract the essential human movement features from the input data to get a prediction of the activity. 

\subsection{Point Cloud-Based HAR}
In vision-based point cloud preprocessing, since vision sensors can generate dense point clouds, downsampling and voxelization methods are commonly used to reduce the data size while keeping essential information. These downsampling algorithms are mostly based on statistical methods. For example, farthest point sampling (FPS) iteratively selects the farthest point from the points already selected previously to ensure the sampled points have a similar statistical distribution with the original points\cite{guo2020deep}. One famous resampling example is the human skeleton estimation algorithm \cite{presti20163d}, which converts over 10,000 points into 20-30 points that represent the key human joints based on the analysis of the human biological structure. In terms of voxelization, it discretizes the 3D space into several regular-size volumetric pixels, known as voxels~\cite{zhou2018voxelnet, gill2011system}. It can reduce the data size by assigning multiple points to one voxel, and the point cloud set becomes a set of uniform-shaped voxels.

Regarding deep learning models, a convolutional neural network (CNN)~\cite{zhou2018voxelnet, liu2019point} is used to efficiently extract the unique spatial information for the voxelized point clouds by using convolutional and pooling layers. Some famous point cloud-based HAR systems like Point Long-Term-Short-Memory (PointLSTM) \cite{min2020efficient} and MeteorNet \cite{liu2019meteornet} are using PointNet \cite{qi2017pointnet, qi2017pointnet++} models, which are specifically designed for processing the point cloud data by individually processing each point in the set and then acquiring global features by using global pool layers. 

The above methods may not be optimal or even not workable for mmWave radar-based HAR because mmWave radar has unique sparsity characteristics. 
For instance, the downsampling strategy used in vision-based methods now needs to be a combination of upsampling and downsampling algorithms. Previous work like \cite{palipana2021pantomime, salami2022tesla, an2021mars} all use clustering-based upsampling algorithms, which bring extra computational efforts but only add some virtual points. The works in~\cite{gong2021mmpoint, cui2024milipoint, an2021mars} use simplified upsampling algorithms like adding zero points or random points, known as zero-padding and random sampling, which bring meaningless points and increase the data redundancy. Besides, voxelization is used in~\cite{singh2019radhar, wang2021m, yu2022noninvasive} to convert the mmWave radar points into uniform-shape cubes. While voxelization is originally designed to reduce the data size in the vision-based point cloud, it results in a lot of empty voxels in the mmWave radar point cloud due to the sparse nature. 

\subsection{Graph-Based HAR}
Besides extracting features from the point cloud, converting the point cloud to a graph is another popular approach~\cite{guo2020deep}. A graph is defined as a set of objects in which some pairs of the objects in this set are connected \cite{wu2022graph}. The graph can build connections between the points based on some practical assumptions, which makes it easier for deep learning models to learn the essential features. Such a graph-based approach has been widely applied in vision-based HAR, namely the skeleton graph \cite{ahmad2021graph, yan2018spatial}. Specifically, a skeleton graph connects the estimated human key joints with biological knowledge. 
Thus, human movements can be efficiently represented by the skeleton graphs in continuous frames~\cite{shi2019skeleton}. 
Their graphical structures can be learned by a graph neural network (GNN)~\cite{wu2022graph} to efficiently extract human-related features from the connections between these points~\cite{yan2018spatial, shi2019skeleton, ahmad2021graph}. 

In mmWave radar-based HAR systems, due to the sparsity and variable size problems, the vision-based skeleton graph construction cannot be directly applied. There are some initial explorations on using graphs for mmWave radar-based HAR, e.g., MMPointGNN \cite{gong2021mmpoint} and Tesla-Rapture \cite{salami2022tesla}. MMPointGNN constructs a graph by connecting all pairs of points in each frame, but it suffers from high computational efforts. Telsa-Rapture constructs a graph by connecting the points across two consecutive frames but does not consider constructing a graph with points in the same frame; it also requires resampling algorithms to keep the same size in consecutive frames. 

The spatial-temporal relationship in the mmWave radar point cloud is not fully explored, which, however, will be essential in human activities because spatial information and temporal information can be used to detect different human body parts and movements of these body parts, respectively.

\sh{
\subsection{Our Contribution}
In this work, we proposed a graph-based HAR system to explore the spatial-temporal features of human movements from the sparse and variable mmWave radar point cloud sequences. We designed a star graph to extract the relative relationship between the radar points and a constant static point,  avoiding the sparsity problem. A discrete dynamic graph neural network (DDGNN) model was proposed to accommodate input graphs of varying sizes and transform them into uniform-sized vectors, solving the challenge of variable-size point clouds without needing additional preprocessing algorithms.
The main contributions of this paper are summarized as follows.

\begin{itemize}
    \item We proposed a star graph structure for mmWave radar-based HAR, where a manually static center point is connected to all radar points. We demonstrate that the relative relationship between a static center point and detected dynamic radar points is crucial in HAR. 
    \item We designed a graph-based HAR pipeline that directly accommodates the sparsity and variable size of mmWave radar data. The proposed DDGNN can process graphs of varying size and generate fixed-length feature vectors for a Bi-LSTM for a temporal feature extractor.
    \item We carried out a comprehensive experiment evaluation based on a real-world collected HAR dataset, which demonstrates that our star graph representation outperforms other baseline methods. Specifically, our directed star graph representation with the DDGNN model achieved an overall test accuracy of 94.27\%. We carefully conducted a practical inference test on the Raspberry Pi platform, showing high inference accuracy with affordable complexity. We also evaluated our DDGGN model with a series of comprehensive ablation studies and recent radar-specific methods. We provided a comprehensive complexity analysis for our system and an extensive discussion on generalizations. 
\end{itemize}
}
The rest of the paper is arranged as follows.
Section~\ref{sec: problem_state} will introduce the problem statement. Section~\ref{sec: system overview} presents the system overview for our graph-based HAR system. Section~\ref{sec: point cloud preprocess} introduces the point cloud preprocessing algorithms used in this work. Section~\ref{sec:graph_generation} discusses the graph generation and introduces \sh{the construction of the star graph sequence.} Section~\ref{sec:DDGNN} presents the general structure of our DDGNN model from the overview of the system to the details of the DDGNN model. Section~\ref{sec: benchmark} introduces the ablation study and the baseline models that we used in the experiments. Section \ref{sec: results} will show the experimental results on our real-world collected dataset on the HAR task. Section \ref{sec: analysis} provides a complexity analysis and discusses the generalizations of our system. Section \ref{sec: related works} summarizes the related works. Finally, Section \ref{sec: conclusion} concludes the paper.

\begin{figure}[!t]
    \centering
    \includegraphics[width=3.4in]{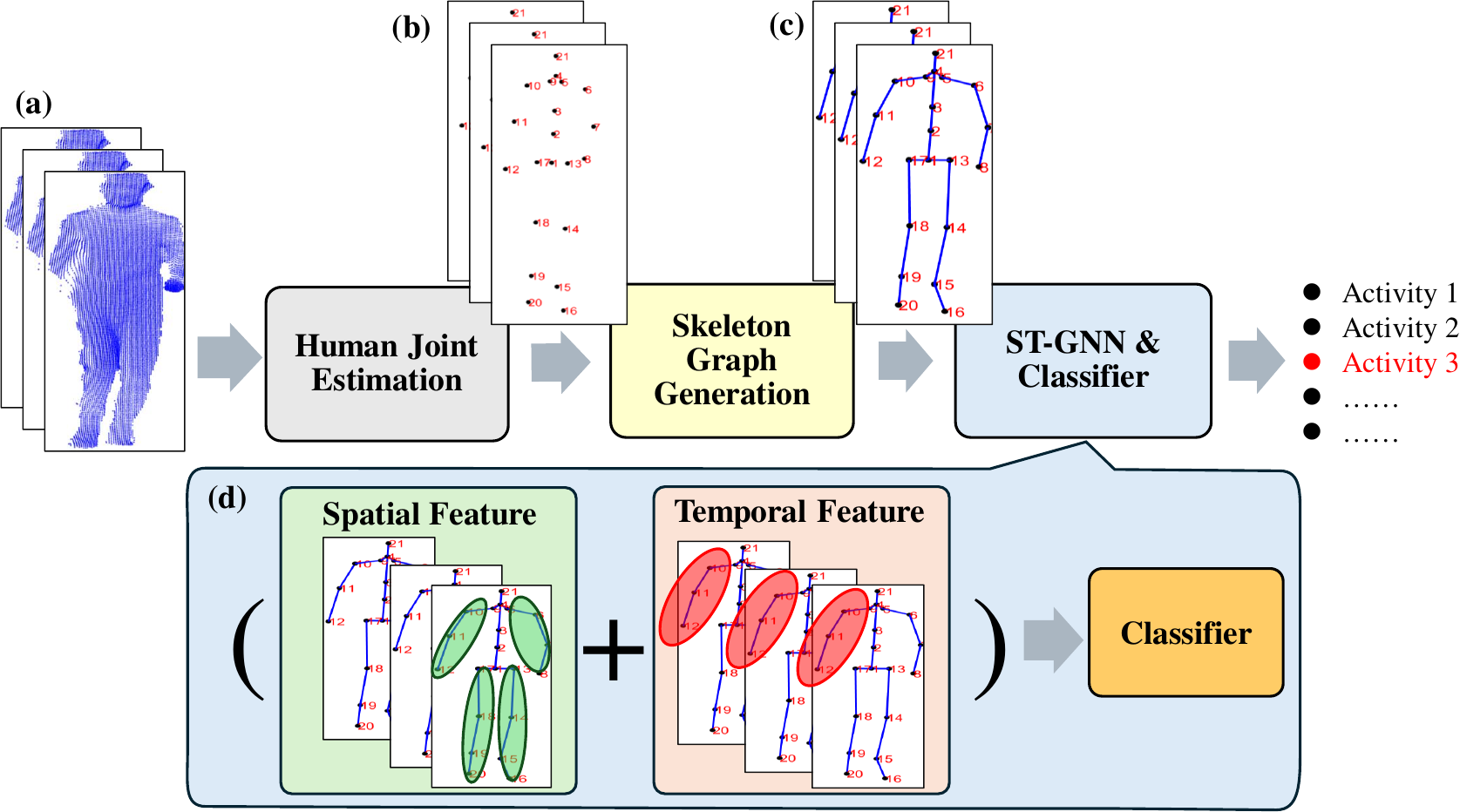}
    \caption{A skeleton graph-based HAR system~\cite{yan2018spatial}.}
    \label{fig: skeleton pipeline}
\end{figure}

\section{Problem Statement}\label{sec: problem_state}

\subsection{Skeleton-Based HAR}

A skeleton graph-based HAR system is depicted in Fig.~\ref{fig: skeleton pipeline}. Fig.~\ref{fig: skeleton pipeline}(a) is the raw Kinect point cloud data captured from the vision-based depth camera Kinect v2. Then, we can use human joint estimation algorithms \cite{shi2019skeleton, cheng2020skeleton} or open-source toolkits, e.g., Kinect body tracking SDK \cite{webb2012beginning}, to obtain the skeleton data shown in Fig.~\ref{fig: skeleton pipeline}(b). Human body information from the raw point cloud data is now converted to a vector that contains the coordinate information of key human body parts. After getting the estimated skeleton data frames, for each frame, these estimated joints are ``connected" by bones (blue lines) to construct a skeleton graph, shown in Fig.~\ref{fig: skeleton pipeline}(c). 
There are spatial features representing the connections between the joints in each skeleton graph, shown as green parts in Fig.~\ref{fig: skeleton pipeline}(d), and temporal features of
the same joints among consecutive graphs, the red parts in Fig.~\ref{fig: skeleton pipeline}(d). Therefore, a spatial-temporal GNN (ST-GNN)~\cite{yan2018spatial,wu2022graph} can be used to explore both the spatial and temporal features. Finally, a classifier is used to generate the predicted results.

By constructing a skeleton sequence with the skeleton data in consecutive frames, the spatial features of the HAR are quantified by analyzing the relative pairwise distance relationship between the joints within the same frame. Meanwhile, temporal features are derived by evaluating changes in the joint locations across consecutive frames \cite{ahmad2021graph}. 

\subsection{Sparsity and Variable Size of mmWave Radar Point Cloud}
\begin{figure}[!t]
\centering
    \includegraphics[height=1.5in]{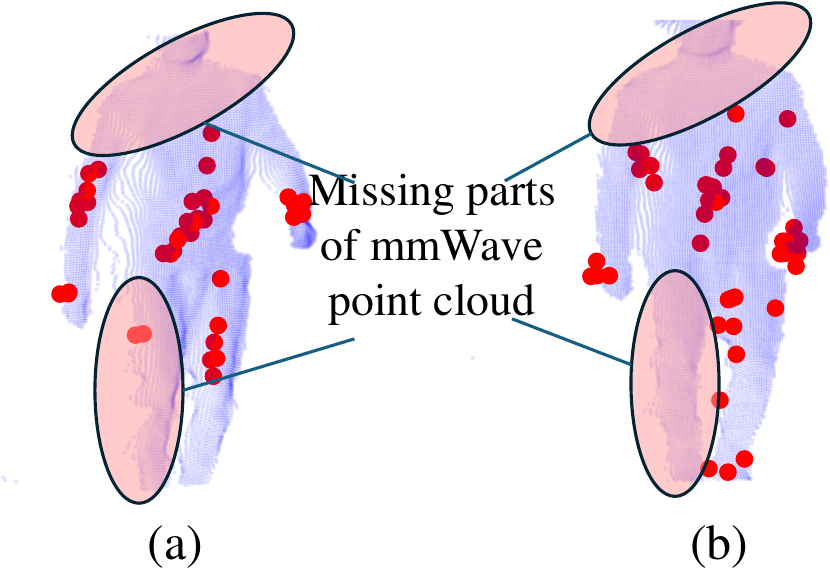}
\caption{Comparisons between the vision-based point cloud and TI mmWave radar point cloud on the same human activity (walking on the spot) at different timestamps (a) $t_1$ and (b) $t_2$. Red points are the TI mmWave radar points, and the blue points are the Kinect raw points (vision-based point cloud).}
\label{fig:kinect_radar}
\end{figure}

Although the TI mmWave radar has several advantages, such as small size and high sensing sensitivity to moving objects, their point clouds are sparse, and the number of points in each frame will be different~\cite{venon2022millimeter, singh2019radhar}. We visualized the TI mmWave radar point cloud and the Kinect point cloud on the same activity in Figs.~\ref{fig:kinect_radar}(a) and \ref{fig:kinect_radar}(b). 

\textbf{Sparsity}: The Kinect point cloud data, marked as the blue points in Figs.~\ref{fig:kinect_radar}(a) and \ref{fig:kinect_radar}(b), shows densely distributed points, over 10,000 points, which enables the construction of complete and stable human body movement across the frames. 
Based on such dense points, we can estimate 21 human joints in Fig.~\ref{fig: skeleton pipeline}(b), connect the joints based on the human body structure knowledge, and finally generate the skeleton graph in Fig.~\ref{fig: skeleton pipeline}(c) from the human joints.
In contrast, a single frame in TI mmWave radar point cloud data ranges from 10 to 50 raw points, which is very sparse. When we want to use the TI mmWave radar point cloud to generate precise skeleton data, the sparse TI mmWave radar point cloud cannot provide complete human body information in 3D space, as it may miss some human body parts, as shown in the marking areas in Fig.~\ref{fig:kinect_radar}. It directly causes the failure of the correct joint estimation for each frame.

\textbf{Variable Size}:  The movement of the human body is not uniform, which may cause the number of moving points to be different across the frames. We consider this data type to be a variable-size sequence. For example, the number of mmWave radar points (marked as red circles) in Figs.~\ref{fig:kinect_radar}(a) and ~\ref{fig:kinect_radar}(b) are 40 and 50, respectively. Such variable-size problems will affect the different spatial distributions in different frames.

The sparsity and variable point numbers cause the pipelines shown in Fig.~\ref{fig: skeleton pipeline} not applicable to the TI mmWave radar point cloud. 
This motivates us to design a new graph representation and a new GNN structure to extract the spatial-temporal features from mmWave point cloud data. 

\section{System Overview} \label{sec: system overview}
\begin{figure}[!t]
    \centering
    \includegraphics[width=3.4in]{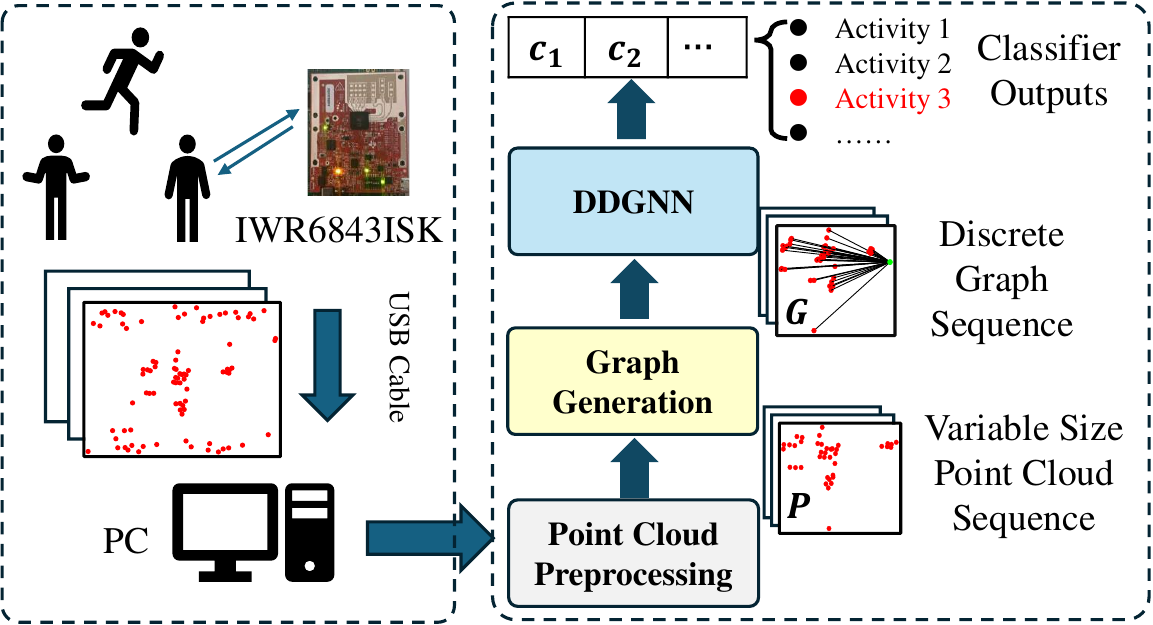}
    \caption{A graph-based mmWave radar HAR system. }
    \label{fig:General Pipeline}
\end{figure}

Our proposed graph-based HAR system with TI mmWave radar is illustrated in Fig. \ref{fig:General Pipeline}, including point cloud preprocessing, graph generation, and the DDGNN model. We will conduct different activities in front of the mmWave radar and use a PC with a USB cable to collect the raw point cloud data using the official TI radar toolbox \cite{ti}. For one single activity, we collect $N$ consecutive frames. 
These $N$ frames of raw point cloud are sent to the preprocessing to remove outliers and noisy points. The preprocessed point cloud sequence $\mathbf{P}$ in each frame is generated to a graph by the graph generation block. 
A sequence of graphs generated from the $N$ frames, $\mathbf{G}$, is input to the deep learning model and the output is the predicted activity class with the highest probability score among all the $C=\{c_1, c_2,...,c_m\}$ classes, representing the $m$ human activities. 

\textbf{Point Cloud Preprocessing:} We perform two preprocessing algorithms to reduce the noise points and store the processed $N$ point cloud frames into a variable-size point cloud sequence $\mathbf{P}$, which will be introduced in Section~\ref{sec: point cloud preprocess}.

\textbf{Graph Generation:} After getting the processed variable-size point cloud sequence $\mathbf{P}$, we generate a star graph for the point cloud frame in each timestamp and store them in a discrete graph sequence $\mathbf{G}$. The proposed star graph focuses on extracting human-related features and addresses the sparsity problem of TI mmWave radar point cloud data. The detailed analysis of star graph representations will be introduced in Section~\ref{sec:graph_generation}.

\textbf{DDGNN:} A DDGNN model extracts the inner-frame spatial features and the cross-frame temporal features from the generated star graph sequence~\cite{wu2020comprehensive, franceschi2019learning}. DDGNN can process the TI mmWave radar point cloud sequence as a sequence of discrete graphs from each frame with the same adjacency matrix, allowing the model to learn spatial and temporal features from the point cloud sequence directly. DDGNN is distinguished from Spatial-Temporal GCN \cite{ahmad2021graph} and Tesla-Rapture \cite{salami2022tesla} since we do not build temporal connections across the frames. 
The proposed DDGNN structure will be introduced in Section~\ref{sec:DDGNN}.

\section{Point Cloud Preprocessing} \label{sec: point cloud preprocess}
To ensure that the collected point cloud data is as accurate as possible to represent the human body movement, two preprocessing algorithms are applied to each frame data as in the works of \cite{cao2022cross} and \cite{wang2023human}. 

Firstly, a noise reduction algorithm is conducted to reduce environmental noise. These environmental noises will cause some unrelated points in each frame, which are considered as the ``ghost points" \cite{cui2024milipoint}. To reduce the impact of these ghost points, we can construct an estimated detection area based on prior knowledge of human heights and the distance between the radar and human locations. In our work, the points where the Z-axis value is larger than 2.5 and smaller than -1 are removed since they are out of the range of human heights. With the estimated distance between the radar and human locations, we can set that the Y-axis values larger than 6.5 and smaller than -1.2 are removed. Also, the X-axis values larger than 5 and smaller than 0.5 are removed. Similar methods of eliminating out-of-range points can also be found in~\cite{salami2022tesla, wang2023human, an2021mars}.

Secondly, a DBSCAN algorithm, which has been used in previous work \cite{wang2024multi, palipana2021pantomime, wang2023human}, is applied to acquire the cluster of points of the human so that we can suppress noise points around the human body. The two major parameters in DBSCAN, the minimum number of points and Epsilon, are set to 2 and 0.35, respectively. We then choose the largest cluster in the DBSCAN output as our output point cloud set. 

After the above preprocessing algorithms, we can obtain the point cloud set at each discrete timestamp, given as $P_n = \{p_1, p_2,..., p_{N_{t_n}}\}$,  where $N_{t_n}$ is the number of points in the $n^{th}$ frame which may vary. $p_n \in \mathbb{R}^3$ represents the 3D location of the detected moving objects. 
For each activity instance, we collect $N$ frames and we store the processed point cloud sets into a sequence as $\mathbf{P} = \{P_1, ...., P_n, ...., P_N\}$.

\section{Graph Generation}\label{sec:graph_generation}
Inspired by the skeleton-based HAR systems, we explore the potential of using star graph representation that efficiently extracts the essential human movement features from the mmWave radar point cloud sequence. In this section, we will first introduce the graph generation from the point cloud, then define the star graph we used in this work, and finally explain the construction method of the star graph. 
\subsection{Graph Generation from Point Cloud}
Define a graph sequence $\mathbf{G} = \{G_1,..., G_N\}$ from a point cloud sequence $\mathbf{P}$ in Section~\ref{sec: point cloud preprocess}, where its element $G_n, n \in(1,N)$ is generated from point cloud set $P_n$ in the frame at the timestamp $n$ with the number of points $N_{t_n}$, expressed as
\begin{equation}\label{eq: graph def}
    G_n = \{P_n, E_{n}\}, P_n \in \mathbf{P}.
\end{equation} 
Each point in the graph may have connections to its neighboring points. We define the connection between two points as the edges stored in the set $E_n$. The number of edges in $E_n$ is denoted as $\mathcal{E}_{t_n}$

\begin{figure}[!t]
    \centering
    \subfloat[]{\includegraphics[height=1.0in]{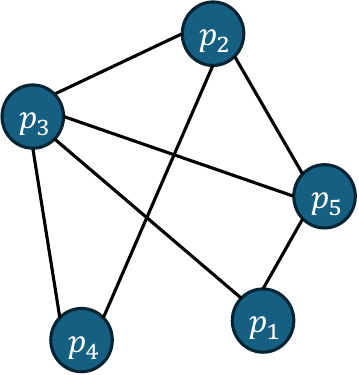}}
     \hspace{0.1in}
    \subfloat[]{\includegraphics[height=1in]{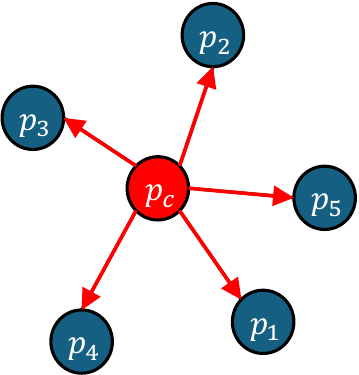}}
     \hspace{0.1in}
    \subfloat[]{\includegraphics[height=1in]{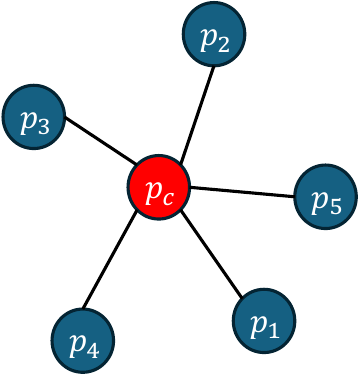}}
    
    \caption{(a) A general graph with 5 points. (b) A directed star graph with the center point $ p_c$; (c) An undirected star graph with the center point $p_c$.}
    \label{fig:graph illu}
\end{figure}

In graph theory, the connection may have directions, which produces an undirected graph and a directed graph. The definition of the neighboring points will be different in the two graphs. The set of the neighbor points stores in $\mathcal{N}(\cdot)$. 
\begin{itemize}
\item Directed graph: The connection from $p_i$ to $p_j$ is defined as $e_{ij}$. We consider $p_i$ to be the neighbor point of $p_j$ (i.e., $p_j \in \mathcal{N}(p_i)$) since the edge originates at $p_i$ and terminates at $p_j$. The set of edges $E_n$ is defined as
    \begin{equation}
        E_n = \{e_{ij} | p_j \in \mathcal{N}(p_i), \forall p_i \in P_n\}.
    \end{equation} 
    \item Undirected graph: The connections between $p_i$ and $p_j$ is defined as $(p_i, p_j)$. Then, $p_i$ and $p_j$ are the neighbor points to each other. The set of edges $E_n$ is defined as
    \begin{equation}
        E_n = \{(p_i, p_j) |  p_j \in \mathcal{N}(p_i), \forall p_i \in P_n\}.
    \end{equation}
\end{itemize}
An undirected graph with 5 points is illustrated in Fig.~\ref{fig:graph illu}(a). For instance, $\{p_3, p_4, p_5 \}$ are connected to $p_2$, which are the neighbor points of $p_2$, given as $\mathcal{N}(p_2) = \{p_3, p_4, p_5 \}$. 

In our work, we will convert the edge set $E_n$ to an adjacency matrix $A_{n}$ as an input of the GCN to represent the connections in the input graph. Thus, (\ref{eq: graph def}) becomes
 \begin{equation}
    G_n = \{P_n, A_n\}, P_n \in \mathbf{P},
\end{equation}
where the adjacency matrix $A_n$ is defined as 
\begin{equation} \label{eq: adj}
A_n(i,j) = 
\begin{cases}
1, & \text{if } p_j \in \mathcal{N}(p_i), \\
0, & \text{otherwise}.
\end{cases}
\end{equation}
$A_n$ is a square matrix with a size of $N_{tn} \times N_{tn} $ that contains all the edges in $E_n$. Specifically, for the adjacency matrix $A_n$ in the undirected graph, when $(p_i, p_j) \in E_n$ exists, means $e_{ji}, e_{ij} \in E_n$. Thus, the $A_n$ of an undirected graph is symmetric, which means $A_n(i, j) = A_n(j, i)$, and its definition can also use (\ref{eq: adj}).

\subsection{Star Graph}
In a star graph, a central point is connected to all other points, but there are no connections between the other points. Consider a central point $p_c$ that is not in the point cloud set \( P_n \). We then define the updated point cloud set as $ P_n' = \{p_c\} \cup P_n$. 

We now define directed and undirected star graphs based on two kinds of connections.

\textbf{Directed star graph (DStar)}: the set of edges is defined as 
    \begin{equation}\label{eq: star_graph_edge_center}
    E_n = \{e_{ic} |  \forall p_i \in P_n', i \neq c \}.
    \end{equation} 
    The corresponding adjacency matrix is given by
    \begin{equation} \label{eq: star_graph_adj}
        A_n(i, j) =
        \begin{cases} 
        1, & \text{when } e_{ic} \in E_n, \\
        0, & \text{otherwise}.
        \end{cases}
    \end{equation}
    
\textbf{Undirected star graph (UStar)}: the set of edges becomes
    \begin{equation}\label{eq: star_graph_edge}
    E_n = \{(p_i, p_c) | \forall p_i \in P_n', i \neq c \}.
    \end{equation}
    In such a definition, $E_n$ has both $e_{ci}$ and $e_{ic}$. The corresponding adjacency matrix $A_n$ is symmetric, expressed as \begin{equation} \label{eq: star_graph_adj_und}
        A_n(i, j) = A_n(j, i) =
        \begin{cases} 
        1, & \text{when } e_{ic} \in E_n, \\
        0, & \text{otherwise}.
        \end{cases}
    \end{equation}
The star graph extracts the center-related information of the point cloud set. We illustrate directed and undirected graphs that have five points and one center point in Fig.~\ref{fig:graph illu}(b) and Fig.~\ref{fig:graph illu}(c), respectively. In the directed graph shown in Fig.~\ref{fig:graph illu}(b), the red connections are 
 $p_c$ pointed to the other points, which have directions. $\mathcal{N}(p_i) = \{p_c\}$ and $\mathcal{N}(p_c) = \emptyset$. 
 In contrast, the connections in the undirected graph shown in Fig.~\ref{fig:graph illu}(c) do not have directions. Then, in Fig.~\ref{fig:graph illu}(c), $\mathcal{N}(p_i) = \{p_c\}$ and $\mathcal{N}(p_c) = \{p_1, p_2, p_3, p_4, p_5\}$.

\subsection{Construction of Star Graph Sequence}
\begin{figure}[!t]
    \centering
    \includegraphics[width=3.4in]{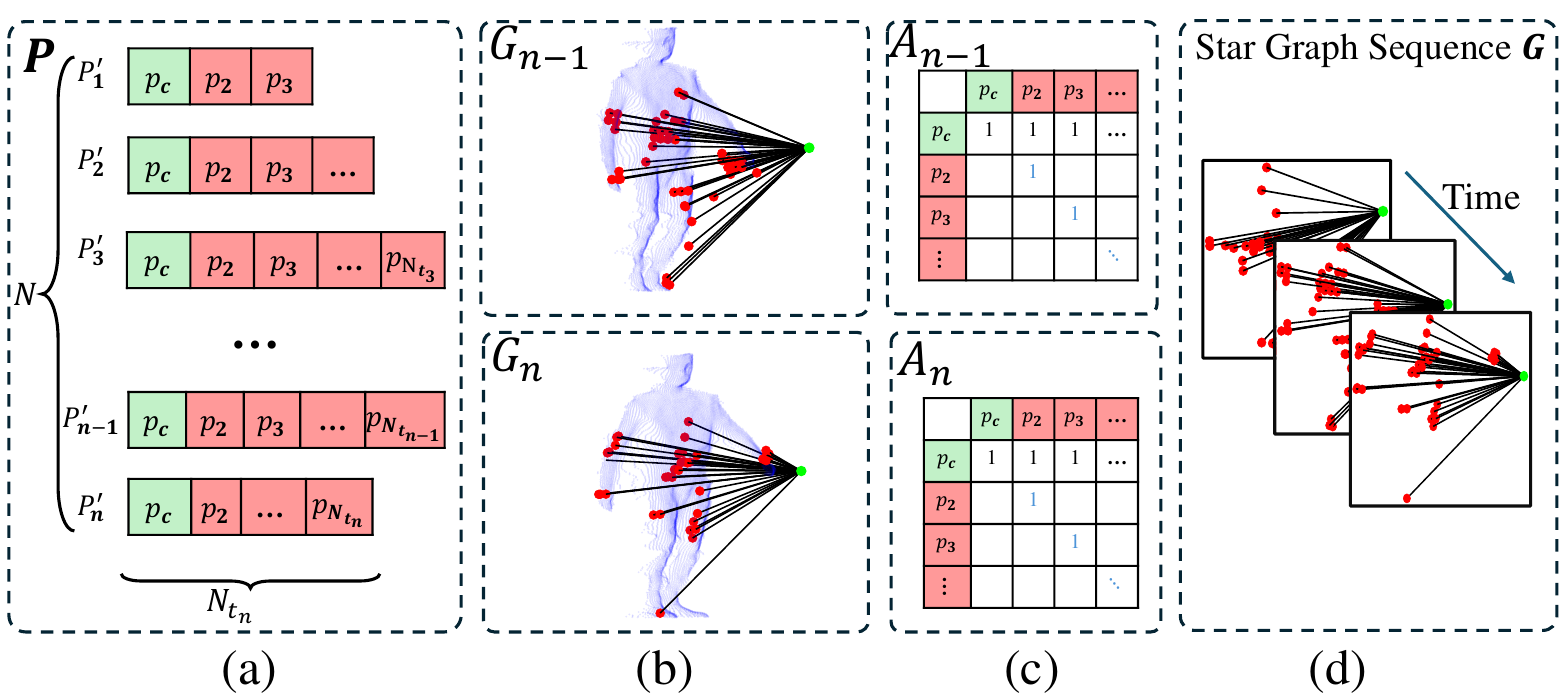}
    \caption{Visualization of star graph construction steps. (a) The variable-size point cloud sequence with manually added static center points. (b) Visualization of the star graph on mmWave radar point cloud at frame $t_{n-1}$ and $t_n$. The red points are the mmWave radar points, and the green point is the center point.  The black lines are the connections in the graph. The blue points are the Kinect point cloud. (c) The illustration of the adjacency matrix of star graphs at frame $t_{n-1}$ and $t_n$. (d) The illustration of the output Star Graph Sequence. }
    \label{fig: physical}
\end{figure}

For each frame in $\mathbf{P}$, the construction steps of the star graph are as follows:
\begin{itemize}
    \item Manually add a constant static center point for each frame as the first point, shown in Fig.~\ref{fig: physical}(a).
    \item Construct a star graph by connecting all the detected radar points to this constant center point. We visualized the star graphs on the two mmWave radar point sets in Fig.~\ref{fig: physical}(b) with $G_{n-1}$ and $G_n$. 
    \item Determine the adjacency matrix by the definition of DStar and UStar, shown in Fig.~\ref{fig: physical}(c). 
\end{itemize}

Then, the output is the star graph sequence $\mathbf{G}$, shown as Fig.~\ref{fig: physical}(d).

\sh{
In the raw mmWave radar point cloud, the static parts are implicit because the TI radar point cloud preprocessing \cite{ti} and our preprocessing algorithms have removed all static objects, including ``ghost points" and environmental noise. The manually added static center point represents the abstracted static part of mmWave radar point cloud, as shown by the green point in Fig.~\ref{fig: physical}(b).

Thanks to the mmWave radar's high sensitivity to moving objects, the preprocessed point cloud set now primarily contains the detected moving objects in the space. When we conduct activities in front of a mmWave radar, the radar detects our body's moving patterns and provides the corresponding 3D location of the moving objects. Therefore, we can consider that most of the preprocessed mmWave radar points can approximately represent the dynamic parts of the human body, as shown by the red points in Fig.~\ref{fig: physical}(b). Different activities cause different parts of the human body to move at various speeds and in different directions, providing unique variable-size point cloud sets in consecutive frames. }

The spatial and temporal features in this star graph sequence are explained as follows:
\begin{itemize}
    \item \textbf{Spatial Features}: Each frame has a different number of points, and their locations are different. It means the ``human moving parts" and their spatial location in each frame are different. In each frame, since we only connect the radar points to the center point, the relative relationship only exists between the single static center point and the other radar points. Thus, the relative relationships stored in each star graph are different, which can represent the different spatial human body features in each frame. 
    \item \textbf{Temporal Features}: In the consecutive frames, the center point is constant and the only dynamic points are the radar points. The temporal features are extracted by evaluating the changes of the relative relationships of the dynamic points and the static center point across consecutive frames. The changing patterns of the dynamic points represent the human body's moving patterns.
\end{itemize}

These spatial and temporal features will be learned by the DDGNN model, explained in Section~\ref{sec:DDGNN}. 

\section{DDGNN}\label{sec:DDGNN}

\subsection{Overview}
\begin{figure}[!t]
    \centering
    \includegraphics[width =3.4in]{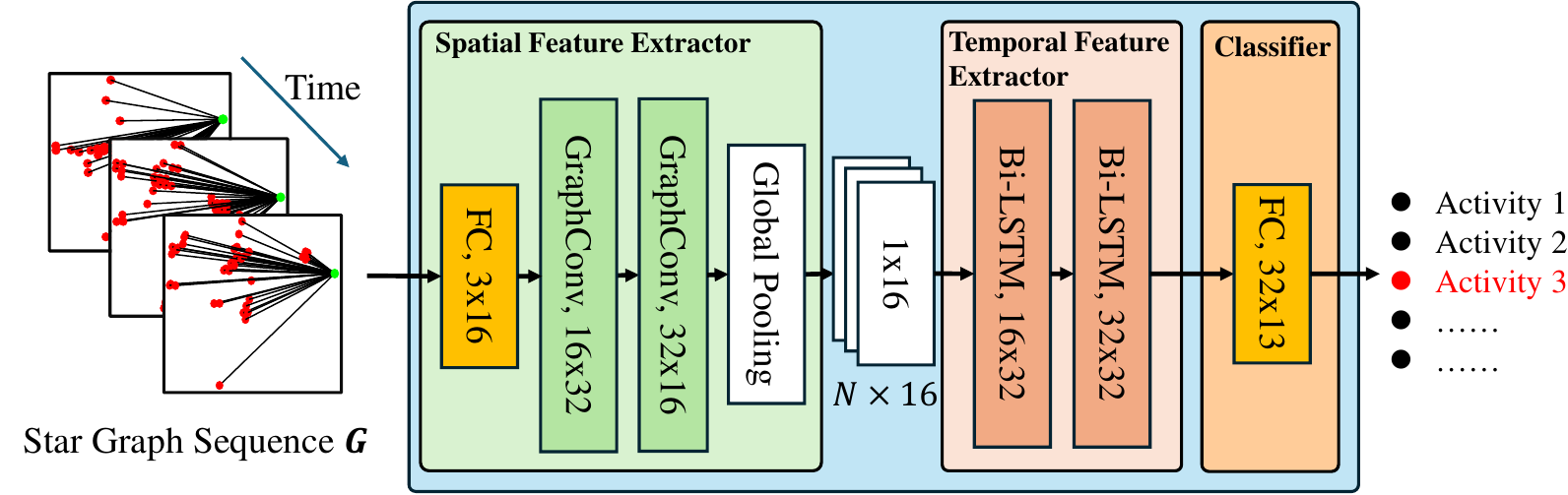}
    \caption{Our proposed DDGNN model.}
    \label{fig: model details}
\end{figure}

Our proposed DDGNN model is shown in Fig.~\ref{fig: model details}, which can be decomposed into three principal components:
\begin{itemize}   
    \item \textbf{Spatial Feature Extractor}: a two-layer GCN with GraphConv~\cite{morris2019weisfeiler} module is employed as the spatial feature extractor to process each graph in the graph sequence $\mathbf{G}$ individually. We then get a $N \times 16$ size vector, where $N$ is the length of the input graph sequence. 
    \item \textbf{Temporal Feature Extractor}: The $N \times 16$ size vector then passes through a two-layer Bi-LSTM model to extract temporal features.
    \item \textbf{Classifier}: A single-layer FC classifier with a softmax function generates the predicted labels. 
\end{itemize}
The goal of our DDGNN model is to explore the spatial-temporal features of human activities from the input graph sequence and recognize them with the extracted features. 

\subsection{Spatial Feature Extractor}
\subsubsection{Structure}
The spatial feature extractor consists of three parts: one FC layer, two GraphConv modules, and a global mean pooling layer. 

Considering the input is $G_n = \{P_n, A_n\}$. It first uses an FC layer to project the 3D coordinates in each point into a higher-dimensional space, expressed as 
\begin{equation}
     H_{i}^{(1)} = \text{FC}(p_i)
\end{equation}
Here,  $p_i \in P_n$. $H_{i}^{(1)} \in \mathbb{R}^{d} $ is the hidden representation for $p_i$. $d$ is the output dimension of the FC layer, also seen as the input dimension of the GraphConv module.

Then, the two-layer GraphConv module processes each graph in the graph sequence. The back-end of the model is PyTorch Geometric~\cite{paszke2019pytorch}. The overview of a single GraphConv module of $l^{th}$ layer can be expressed as
\begin{equation}
    H^{(l+1)} = \text{GraphConv}(H^{(l)}, A_n),
\end{equation}
where $H^{(l)}$ represents the $l^{th}$ layer representation and $A_n$ is the adjacency matrix. Since we have known the relationship of $A_n$ and $\mathcal{N}(p_i)$ in (\ref{eq: adj}), the generalized form of the $l^{th}$ layer of the GraphConv is expressed as
\begin{equation} \label{eq: graphconv 2}
H^{(l+1)}_i = \sigma\Big((W_1^{(l)}+W_2^{(l)})H^{(l)}_i + W_2^{(l)}\bigoplus_{j \in \mathcal{N}(p_i)} (e_{ij}, H^{(l)}_j)\Big),
\end{equation}
where $\sigma(\cdot)$ denotes the sigmoid activation function, $H_i^{(l)} $ is the representation of $p_i$ at the $l^{th}$ layer.  $W^{(l)}_1 $ and $W^{(l)}_2$ represent the linear transformation on each point representation at the $l^{th}$ layer. $\bigoplus$ represents the aggregation rule such as add, max, or mean. In this work, we use ``add" as our aggregation rule.

For clarity, the layer weights use the standard form $W_1^{(l)}, W_2^{(l)} \in \mathbb{R}^{F_{\mathrm{in}}^{(l)} \times F_{\mathrm{out}}^{(l)}}$, so that $H^{(l)} \in \mathbb{R}^{N_{t_n} \times F_{\mathrm{in}}^{(l)}}$ is mapped to $H^{(l+1)} \in \mathbb{R}^{N_{t_n} \times F_{\mathrm{out}}^{(l)}}$ after aggregation and transformation. $F_{\mathrm{in}}^{(l)}$ and $F_{\mathrm{out}}^{(l)}$ represent the input and output dimensions of the $l^{th}$ layer GraphConv module, respectively.

Note that $W^{(l)}_2$ exists in the first part since we allow the GCN to consider the point itself as a part of neighbor points. Such a process is called self-loop. Since the point in the graph has its own coordinate information, the self-loop can stabilize the representation learning at each layer. 

\begin{figure}[!t]
    \centering
    \includegraphics[width = 3.0in]{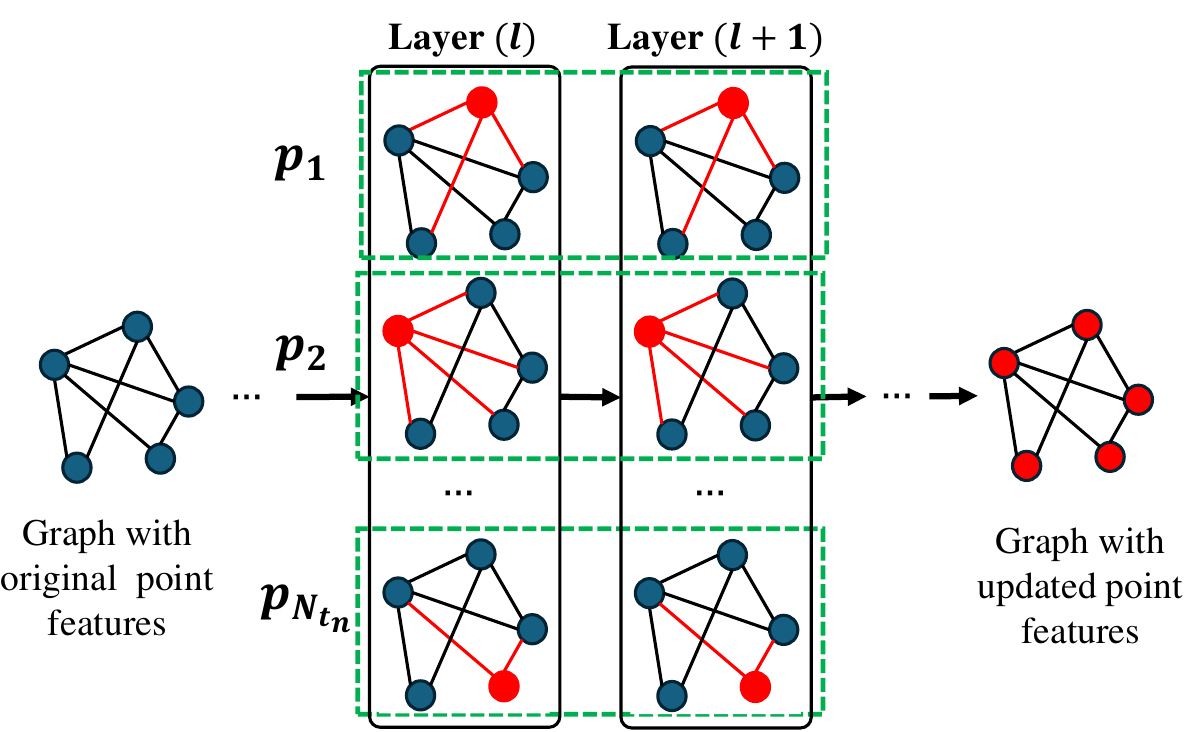} 
    \caption{An illustration of GraphConv module information aggregation rules with a general graph with $N_{tn} = 5$. GraphConv module will iteratively aggregate feature information from each point's neighbors (shown in red points).}
    \label{fig:GCN model}
\end{figure}

An illustration of the GraphConv module on a general graph with $N_{t_n} = 5$, same as Fig.~\ref{fig:graph illu}(a), is shown in Fig.~\ref{fig:GCN model}. The process of aggregation of information shown in (\ref{eq: graphconv 2}) will be iterated to all points at each layer, marked as the changes of the red points in Fig.~\ref{fig:GCN model}. The output graph maintains the same graph structure but has updated the features of all points. Thus, the aggregated information of GCN is completely based on the connections in the graph,  which is more flexible than the CNN and PointNet structures. 

After the two-layer GraphConv module, we used the dropout layer to avoid overfitting and increase the representation generalization ability. Then, we choose to use the global mean pooling layer to aggregate the essential information of all points into one representation vector, thus, owning in the same dimension of the output of the GraphConv module, $V_n \in \mathbb{R}^{1 \times F_{\mathrm{out}}^{(2)}} $. In our work,  the size of $V_n$ is $1 \times 16$. This is the same as the popular vision-based point cloud methodologies~\cite{qi2017pointnet, qi2017pointnet++, guo2023point}.

We define the output representation of $i^{th}$ point of the second GraphConv module as $H^{(out)}_i$. The expression of global meaning pooling is 
\begin{equation}
    V_n = \frac{1}{N_{tn}}\sum_{i=1}^{N_{t_n}} H^{(out)}_{i}.
\end{equation}

The output sequence of the spatial feature extractor is $\mathbf{V}  = \{V_1,..., V_N\} \in \mathbb{R}^{N \times F_{\mathrm{out}}^{(2)}}$. Thus, in this way, the input point cloud sequences with variable sizes are converted to the output representation sequences with the same size. This representation sequence will then be fed into the Bi-LSTM model for the temporal feature extraction.

\subsubsection{Representation Learning on Star Graph} \label{sec:rep star}
(\ref{eq: graphconv 2}) reveals that the GraphConv aims to learn the representation of each point $p_i$ by aggregating the information from its neighbor set $\mathcal{N}(p_i)$. Thus, the representation learning in GraphConv module can be explained as learning the self-point representation learned by $W_1^{(l)}+W_2^{(l)}$ and the aggregated neighbour representation learned by $W_2^{(l)}$. 

In this work, we have directed and undirected star graphs. Their definition of $\mathcal{N}(\cdot)$ is different. Thus, it is necessary to analyze the representation learning of these two star graphs. 

\textbf{(a) Representation Learning on DStar.}\label{sec: rep on dstar}
According to (\ref{eq: star_graph_edge_center}), $\forall p_i \in P_n$, $\mathcal{N}(p_i) = p_c$. When the GraphConv module is applied to $p_i$, (\ref{eq: graphconv 2}) becomes \begin{equation}\label{eq. graphconv_star_i}
H^{(l+1)}_i =  \sigma\left((W_1^{(l)} + W_2^{(l)})H^{(l)}_i+ W_2^{(l)}H^{(l)}_c \right),
\end{equation}
which shows that for any $p_i \in P_n$  in the star graphs, the representation of $p_i$ is learned by integrating the information between the points and the center-point information $W_2^{(l)}H^{(l)}_c $.

Meanwhile, for $p_c$, 
since $\mathcal{N}(p_c) = \emptyset$ and we allow self-loop in GCN model,  (\ref{eq: graphconv 2}) becomes \begin{equation}\label{eq. graphconv_star_i_2}
H^{(l+1)}_c =  \sigma\big((W_1^{(l)} + W_2^{(l)})H^{(l)}_c\big),
\end{equation}
which means that for $p_c$ in the directed star graph, the center point's representation is only learned by $(W_1^{(l)} + W_2^{(l)})$. 

In a two-layer GCN model, the representation learning at the second layer will be affected by the first layer. Thus, for $p_i$ at second layer, combine (\ref{eq. graphconv_star_i}) and (\ref{eq. graphconv_star_i_2}), we can get 
\begin{align} \label{eq: dstar_i2}
    H^{(out)}_i = & \sigma \Big((W_1^{(2)} + W_2^{(2)})H^{(2)}_i +\nonumber \\ & W_2^{(2)}\sigma\big(W_1^{(1)} + W_2^{(1)})p_c \big) \Big),
\end{align} 
which shows that the $p_i$ at the second layer will be affected by $p_c$ at the first layer.  

\textbf{(b) Representation Learning on UStar.}\label{sec: rep on ustar}
According to(\ref{eq: star_graph_edge}), $\forall p_i \in P_n$, $\mathcal{N}(p_i) = p_c$. Thus, the representation of $p_i$ in the UStar is the same as (\ref{eq. graphconv_star_i}).

However, since we consider connections to the $p_c$ in the undirected star graph, $\mathcal{N}(p_c) = P_n$. Then, (\ref{eq: graphconv 2}) for $p_c$ becomes,
\begin{equation}\label{eq: graphcon_c_undirected}
    H^{(l+1)}_c = \sigma\Big( (W_1^{(l)} + W_2^{(l)})H^{(l)}_c + W_2^{(l)} \sum_{j=1}^{N_{t_n}} H^{(l)}_j \Big),
\end{equation}
which shows the representation of the center point at $l^{th}$ layer is learned by aggregating information from all other points in $P_n$ at each layer. This is different from (\ref{eq. graphconv_star_i_2}) in the directed star graph. 

For representation learning of $p_i$ in the UStar at a two-layer GCN model, combining (\ref{eq. graphconv_star_i}) and 
(\ref{eq: graphcon_c_undirected}), we can get
\begin{align}\label{eq: graphconv_propaback}
H^{(out)}_i = & \sigma\Big((W_1^{(2)} + W_2^{(2)})H^{(2)}_i + \nonumber \\  & W_2^{(2)}\sigma\big( W_1^{(1)}p_c + W_2^{(1)} \sum_{j=1}^{N_{t_n}} p_j \big) \Big),
\end{align}
which means that the representation learning on $p_i$ at the second layer will be affected by the other points' information at the first layer, referring to the item $W_2^{(1)} \sum_{j=1}^{N_{t_n}} p_j$. 

Comparing (\ref{eq: graphconv_propaback}) with (\ref{eq: dstar_i2}), we find that the output representation of $p_i$ learn different information from DStar and UStar. We will evaluate their impacts with the experiments in Section~\ref{sec: results}.

\subsection{Temporal Feature Extractor}
Temporal features are essential for human activity recognition because they describe movement patterns over time. Conceptually, the GCN concentrates spatial relations (center–point structure) with a global pooling layer into $V_n$, while the Bi-LSTM captures how these relations evolve over the $N$ frames with the output sequence of the spatial feature extractor $\mathbf{V}$. 

Therefore, we apply a two-layer Bi-LSTM \cite{sak2014long} to extract temporal features
\begin{equation}
\label{eq:bilstm_re}
V_{\text{temporal}} = \mathrm{Bi\mbox{-}LSTM}(\mathbf{V}),
\end{equation}
where $V_{\text{temporal}} \in \mathbb{R}^{1 \times F_{\mathrm{LSTM}}}$. $F_{\mathrm{LSTM}}$ is the output dimension of the Bi-LSTM module.
A Bi-LSTM is a natural temporal head for such 1D sequences, as it augments a recurrent unit with a memory cell, which enables the model to capture long-range dependencies and mitigate the vanishing gradient problem. The bidirectional structure allows the model to capture both forward and backward temporal dependencies. This standard mechanism is well-established in sequence learning (such as LSTM and bidirectional RNNs) and has been widely used as a temporal feature extractor in mmWave radar HAR tasks \cite{singh2019radhar,yu2022noninvasive, shrestha2020continuous}. The two-layer architecture enhances the model's capacity to capture complex temporal patterns and improves its learning ability.

In general, any deep learning model with temporal modeling capability can be considered as a potential choice for the temporal feature extractor. However, considering the trade-off between computational complexity and accuracy in the mmWave point cloud scenario, we choose the Bi-LSTM model as our temporal feature extractor. The comparison of different temporal feature extractor choices will be presented in Section~\ref{sec: results}.

\subsection{Classifier}
Finally, a single-layer FC classifier is used for the classification. The cross-entropy loss \cite{paszke2019pytorch} is used, defined as
\begin{equation}
\mathcal{L} = -\frac{1}{N_{total}} \sum_{i=1}^{N_{total}} \sum_{c \in C} \mathbf{1}\{y_i == c\} \log \hat{y}_{ic},
\end{equation}
where $N_{total}$ is the number of the samples, $\mathbf{1}(\cdot)$ denotes the indicator function, which equals 1 if the condition inside is true, and 0 otherwise, $y_i$ is the true class label for the $i^{th}$ sample,  $\hat{y}_{ic}$ is the predicted probability of the $i^{th}$ sample belonging to class $c$, which is obtained by applying a softmax activation function on the raw output vector $\mathbf{V}_{out}$ from the model.

\section{Baseline Approaches} \label{sec: benchmark}
In our work, we use two types of baseline approaches for comparison: a) Graph-based Baselines: We generate different graph types to replace the star graph in the ``Graph Generation" and keep the DDGNN model the same; b) Non-Graph-based Baselines: We made comparisons with popular point cloud representations using different baseline models. We replaced the ``Graph Generation" block and the ``GCN" model in graph-based HAR systems with these baseline point cloud processing algorithms. Note that in these two methods, the information on the center point is not used since they extract human-related information from the graph or directly from the point cloud. 

\subsection{Graph-based Baselines}
\textbf{Vision-based Skeleton Graph (Skeleton)}: The skeleton graph is generated from the Kinect point cloud. We used the estimated human key points from Kinect as a reference and generated a 25-point skeleton graph for each frame. The performance of the skeleton graph can be used to assess the upper limit of the system, as it provides the most accurate estimation of the human joints. 

\textbf{K-Nearest Neighbor graph (KNN)}: 
KNN graph extracts spatial features by calculating the pairwise Euclidean distance in the point cloud set and choosing the k-nearest neighbors for each point. This means each point will need to calculate the Euclidean distance with all other points in this set.
Given a $k > 0$, the edges of the KNN graph are defined as
\begin{equation}
E_n = \{e_{ij} | p_i \in \mathcal{N}_{knn}(p_j), \forall p_i \in P_n\}.
\end{equation}
where $\mathcal{N}_{knn}(p_j)$ denotes the set of $k$-nearest neighbors of $p_j$. 
The corresponding adjacency matrix is given by
\begin{equation}
    A_n(i,j)  = 
    \begin{cases} 
    1 & \text{if } i \in \mathcal{N}_{knn}(p_j), \\
    0 & \text{otherwise}.
    \end{cases}
\end{equation}
The KNN graph was used in Dynamic Graph Convolutional Neural Network (DGCNN) as an efficient feature extractor \cite{wang2019dynamic}, and several works mentioned its effectiveness in GCNs \cite{li2021towards}. However, the choice of $k$ is crucial, as a small $k$ may not capture sufficient neighbor information, while a large $k$ may introduce noisy connections. Some frames might not have enough points to build up a KNN graph based on the parameter $k$. In this work, since each frame of our mmWave radar point cloud is between about 5 from 50, based on the results in \cite{wang2019dynamic}, when $k=5$, the proposed network has reached the near-optimal performance, we set $k=5$ since our mmWave radar point cloud is much sparse than the vision-based point cloud. 

\textbf{Radius graph (Radius)}: 
The radius-based nearest neighbors are chosen by limiting the connections to a specific radius based on the calculation of the pairwise Euclidean distance in the point cloud set, usually considered as a ball-query information in vision-based point cloud tasks \cite{qi2017pointnet++}. 
Given a radius $r > 0$, the edge set $E_n$ of a radius graph is defined as
\begin{equation} 
E_n = \{e_{ij} | p_i \in \mathcal{N}_{radius}(p_j), \forall p_i \in P_n\},
\end{equation} 
where $\mathcal{N}_{radius}(p_i)$ represents the radius-based nearest neighbor set of the point $p_i$, is defined as
\begin{equation}
    \mathcal{N}_{radius}(p_i) = \{p_j | d(p_i, p_j) \leq r, \forall p_i \in P_n \},
\end{equation}
where $d(p_i, p_j)$ is the distance between points $p_i$ and $p_j$. Define the number of edges in the radius graph as $M \in [0, N_{t_n}^2]$. The corresponding adjacency matrix is given by
\begin{equation}
    A_n(i,j)  = A_n(j, i)  = 
    \begin{cases} 
    1 & \text{if } i \in \mathcal{N}_{radius}(p_j), \\
    0 & \text{otherwise}.
    \end{cases}
\end{equation}
Note that the chosen radius value is essential, similar to the KNN graph, a small radius value may cause less or even no connection among point pairs, while a large radius value may lose some detailed neighbor information. The default values of $r$ used in \cite{qi2017pointnet++} are 0.2, 0.4. Since our mmWave radar point cloud is much sparser than the vision-based point cloud, we set $r=0.5$ to ensure the radius graph can properly be generated in most of the frames.

\textbf{Fully-Connected Graph (FC)}: 
The FC graph provides a dense representation of the relationships among all points in the mmWave radar point cloud. By considering connections between all pairs of points, an FC graph may capture global dependencies and interactions of each frame. 
For each point $p_i \in P_n$, the neighbor set of each point is $\mathcal{N}(p_i) = P_n$. The edges of the FC graph are defined as
\begin{equation}
E_n = \{e_{ij} | \forall p_i, p_j \in P_n\}.
\end{equation}
The corresponding adjacency matrix is given by

\begin{equation}
    A_n(i, j)  = 
    \begin{cases} 
    1 & \text{if } j \neq i, \\
    0 & \text{otherwise}. \\
    \end{cases}
\end{equation}

However, high connectivity can introduce high computational complexity. The model needs to learn to filter out irrelevant information if necessary. 

\textbf{Empty Graph (Empty)}:
    An empty graph refers to a graph without edges. We define its set of edges as $ E = \emptyset$ because the neighbor set of each point is $\mathcal{N}(p_i) = \emptyset$. And the corresponding adjacency matrix is given by $ A_n(i,j)  = \textbf{0}^{N_{t_n} \times N_{t_n}}$, which means that $A_n$ is an all-zero matrix. 
    All points of an empty graph do not share any information, and there are only linear transformations on each point. Such a special case is essential since it can verify whether only the relative relationship in the star graph can provide a positive impact on human-related feature extractions.

\subsection{Non-Graph-based Baseline}\label{sec:nongraph}
We also studied non graph-based inputs.
Four deep learning approaches and one non-deep learning approach are used for baselines.

\textbf{MMPointGNN} \cite{gong2021mmpoint}: MMPointGNN is a specialized PointGNN model for the mmWave radar point cloud-based HAR. It modifies the inner structure of the original vision-based PointGNN model so that the model can learn the essential connections among the points in each frame. It uses the zero padding point sequence as the preprocessing and automatically learns a suitable graph structure from the input point cloud. This is to evaluate the performance of the current work with specialized GNN models on mmWave radar point cloud data. 

\textbf{Voxelization-3DCNN (Voxel)}: Based on Radhar \cite{singh2019radhar}, we designed a voxelization algorithm to convert the 3D point sequence into high-dimensional voxel sequence ($(N_{t_n} \times 10 \times 32 \times 32)$) and used a two-layer 3D-CNN model to process the voxel sequence. This can be used to evaluate the conventional CNN on voxelized mmWave radar point cloud data. Each convolution has a kernel of 3 and padding of 1. 

\textbf{Resampled Point Cloud-PointNet (RS)}: Based on previous work \cite{palipana2021pantomime, salami2022tesla}, we chose the FPS algorithm as the downsampling algorithm and the agglomerative hierarchical clustering algorithm as the upsampling algorithm. The sample number $N_s$=32 is the same as the setting in \cite{wang2023human}. PointNet \cite{qi2017pointnet} uses two linear layers with the same dimension as 3DCNN and the GraphConv module. PointNet individually processes each point and then uses a global mean pooling layer as the output feature.
    
\textbf{Zero Padded Point Cloud-PointNet (ZP)}: Zero padding means it will add zero points if the number of points in this frame is smaller than the expected sample number $N_s$. We set sample number $N_s$ = 32 to keep the same dimension with RS. If the number of points in the frame is larger than 32, we use only the first 32 points. The PointNet model was also the same as the RS settings.

We also did a comparison with a non-deep learning approach.

 \textbf{Statistical features-Logistic Regression Classifier (LR)}: For each sample, we generated the feature sequence with five chosen statistical features: "Mean Value", "Median Value", "Standard Deviation", "Minimal Value", and "Maximum Value". Then, an LR classifier was used.

\section{Experiment Evaluation} \label{sec: results}
\subsection{Experiment Settings}
\subsubsection{Dataset Detail}
\begin{figure}[!t]
    \centering
    \includegraphics[width=0.8\linewidth]{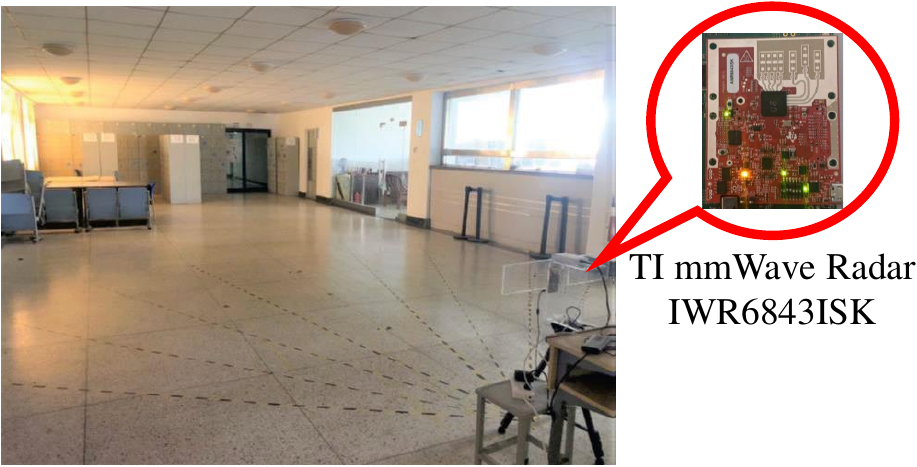}
    \caption{A photo of the collection environment and mmWave radar.}
    \label{fig:environment}

\end{figure}

The datasets used in this paper were collected in an empty indoor environment to avoid any obstacles that could cause strong reflections of radar signals. The radar used was TI-IWR6843ISK, operating in a frequency band ranging from 60~GHz to 64~GHz. The detection frequency was 15 frames per second. This device is equipped with three transmitting antennas (Tx) and four receiving antennas (Rx), which together form a 60-degree field of view (FoV) in both the azimuth and elevation planes, with an angular resolution of 15 degrees. The radar was equipped at a height of 1 meter. The distance between the human target and the radar device was 3 to 5 meters. A photo of the dataset collection environment and the device's location is shown in Fig.~\ref{fig:environment}. 

During data collection, five participants were asked to repeat $m=$ 13 human activities several times. These activities include 1) walking on the spot, 2) rotating the body, 3) clapping, 4) swinging two arms upward and downward, 5) swinging two arms horizontally, 6) lifting the left arm up and down, 7) lifting the right arm horizontally, 8) lunging the left leg, 9) lunging the right leg, 10) walking back and forth, 11) walking back and forth with arm motion, 12) walking in a clockwise pattern, and 13) walking in a counterclockwise pattern. The data collection was approved by the authors' institution. The total collected point cloud data reaches 197,550 frames. Each data sample contained a complete activity from start to end with $N=$50 frames. Each person had 60-70 samples per class.

The data from the first three participants were set as the training set, and the data from the fourth and fifth participants were set as the validation set and the test set, respectively. All the experiment results are plotted using the test dataset for fair comparisons.

\subsubsection{DDGNN Hyperparameter Settings}
The dimension of the DDGNN model is shown in Fig.~\ref{fig: model details}. A constant star center location was set as (0, 1, 0). The dropout rate is set to 0.3. The training loss is the cross-entropy loss, and GCN, Bi-LSTM, and FC classifier parameters were updated simultaneously. The optimizer was Adam. The learning rate was set to 0.001. Due to the variable-size input graphs, during training, we do not use them with batch training strategy. We set the validity of the accuracy after each of the five training epochs was checked to avoid overfitting. An early stop was used to check whether the validation accuracy did not improve in the next 10 epochs. 

\subsection{Classification Accuracy Performance}
\label{sec: graph representation results}
\begin{table}[!t]
\centering
\caption{Overall Test Accuracy }
\begin{tabular}{llc}
\hline
& \textbf{Name} & \textbf{Overall Accuracy (\%)} \\
\hline

\textbf{Ours} & DStar & 94.27 \\
 & UStar & 88.79 \\
\hline

 & Skeleton & 97.25 \\
 & KNN & 90.76 \\
\textbf{Graph-based} & Radius & 86.92 \\
\textbf{Baselines} & FC & 56.00 \\
 & Empty & 88.03 \\
\hline

 & MMPointGNN & 90.58 \\
 & Voxel & 79.85 \\
\textbf{Non-graph-based} & RS & 80.65 \\
\textbf{Baselines} & ZP &  78.95 \\
 & LR & 51.37 \\
\hline
\end{tabular}
\label{tab:overall_accuracy}
\end{table}
\begin{figure*}[!t] 
    \includegraphics[width=7in]{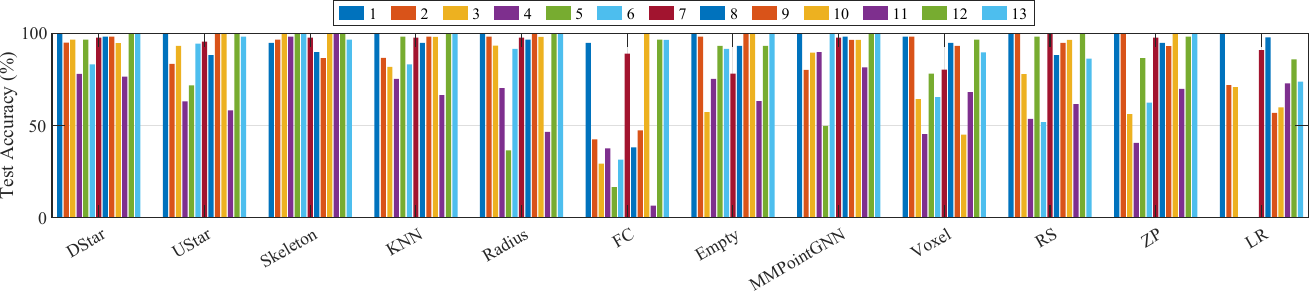}  
    \caption{Class specific accuracy among different baseline models.}
    \label{fig:class_accuracy}
\end{figure*}
We investigated the effectiveness of various representations and baseline methods for the HAR task using mmWave radar point clouds. Table~\ref{tab:overall_accuracy} presents the overall test accuracies of three parts: a) our star graph with DDGNN models; b) other graph types with DDGNN models as ablation study; c) Different baseline models. 
Fig.~\ref{fig:class_accuracy} shows the class-specific accuracy for each approach, which allows us to evaluate the precise feature extraction capabilities of different methods.

From Table~\ref{tab:overall_accuracy}, we can find that DStar has the highest test accuracy (94.27\%) and reaches the near-optimal performance with a skeleton graph (97.25\%), which proves the effectiveness of the proposed method in extracting efficient human-related features. Dstar and Skeleton experiments have no low-accuracy classes, and the accuracies of all classes are higher than 70\%. UStar only gets a test accuracy of 88.79\%. In a UStar graph, the direction is lost, which causes the loss of essential information.

\begin{figure}[!t]
    \centering
    \includegraphics[width=\linewidth]{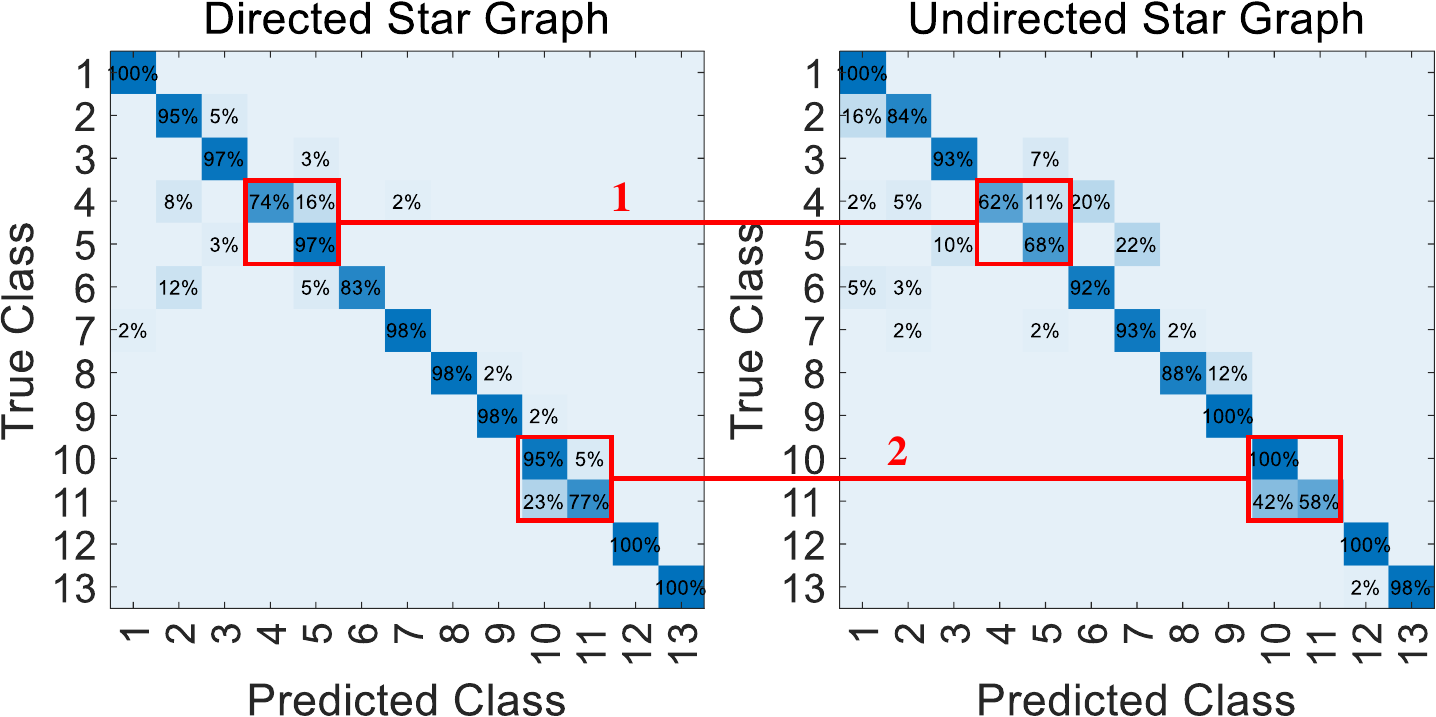}
    \caption{Confusion matrices of DStar and UStar. Red part 1 shows the classification accuracy comparison between two similar classes: 4) swinging two arms upward and downward, and 5) swinging two arms horizontally. Red part 2 shows similar classes: 10) walking back and forth, 11) walking back and forth with arm motion. }
    \label{fig:conf_comparison}
\end{figure}

We visualized the confusion matrix comparison of UStar and DStar in Fig.~\ref{fig:conf_comparison}, which shows class pairs 1 and 2 are complicated classes that DStar performs better than UStar. To solve this issue, a data augmentation strategy \cite{salami2022tesla, palipana2021pantomime, yu2022noninvasive} can be employed to increase the dataset's diversity to enhance the model's robustness. Furthermore, the multi-modal sensing with vision-based sensor and mmWave radar can help the mmWave radar HAR system be robust to the complicated activities \cite{wang2024multi}. Another potential method is to use the hierarchical classification to firstly recognize the simple activities and then extract fine-grained information for complicated activities, which is common in vision-based HAR \cite{shao2020finegym}.

\subsubsection{Comparison with Graph-based Baseline Approaches}
\begin{figure}[!t]
    \centering
    \includegraphics[width=3.4in]{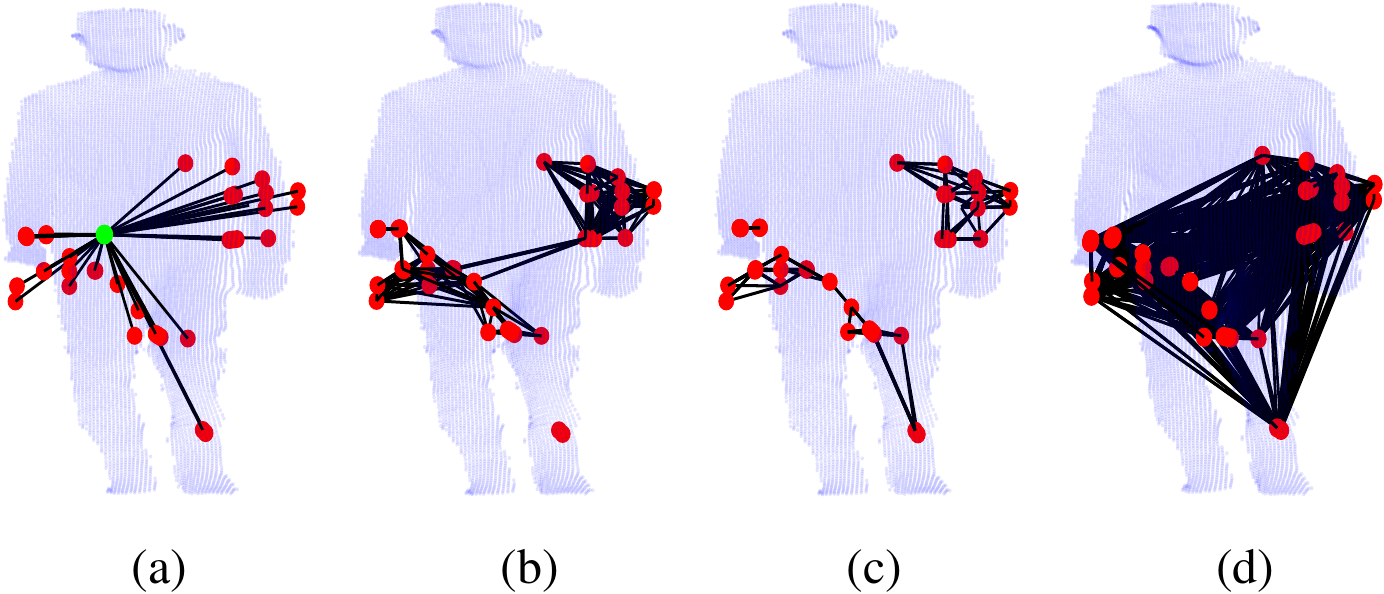}
    \caption{Visualization of different graph representations on the same mmWave radar point cloud frame: (a) Star graph; (b) Radius graph; (c) KNN graph; (d) FC graph.}
    \label{fig:visual_of_graphs}
\end{figure}

\sh{To better explain the different topologies of different graphs, we visualized the four different graphs with mmWave radar points (marked as red) and Kinect points (marked as blue) in Fig.~\ref{fig:visual_of_graphs}. UStar and DStar share the exact visualization, shown in Fig.~\ref{fig:visual_of_graphs}(a). Thus, the star graph captures the relative relationships between the abstracted static part of mmWave radar point cloud (the green center point) and dynamic parts of the human body (the red points). 

An empty graph to get 88.03\% test accuracy, as shown in Table~\ref{tab:overall_accuracy}. However, the empty graph has only 57.38\% and 63.33\% accuracies on classes 3 and 11, as can be observed in Fig.~\ref{fig:class_accuracy}. This proves that the empty graph, without any connections, causes the loss of neighbors in (\ref{eq: graphconv 2}), leading to only the linear transformation of each point independently. This makes GCN difficult to extract precise spatial features from the mmWave radar point cloud, as it loses all neighbors in each frame.

KNN graph and Radius graph have a similar procedure of using pairwise distance calculation to choose the neighbor's points and get accuracies of 90.76\% and 86.92\%, respectively. Looking at the class-specific accuracy, the radius graph gets low accuracies on classes (4, 5, 11) with 70.49\%, 36.67\%, and 46.67\%, respectively. KNN graph gets low accuracies on classes 4 (75.41\%) and 11 (66.67\%). 

As we discussed before, distance-based neighbor search algorithms highly depend on the key parameter choice ($k$ and $r$). If the key parameter is properly chosen, it will generate a regionally connected graph for most of points in Fig.~\ref{fig:visual_of_graphs}(b). Otherwise, if the key parameter is too small, some frames might generate two subgraphs, as shown in Fig.~\ref{fig:visual_of_graphs}(c). In such a situation, the information in the two small subgraphs cannot be shared when passing to the GCN module. Due to the sparse and variable-size nature of the mmWave radar point cloud, the distance-based neighbor search algorithms in KNN and Radius graphs cannot avoid the subgraph issues. This may cause the representation learning in (\ref{eq: graphconv 2}) to be separated into different subgraphs in some frames, which leads to the accuracy drop on complicated classes. 

On the contrary, if the key parameter is too large, the distance-based graphs like KNN or Radius graphs become an FC graph like Fig.~\ref{fig:visual_of_graphs}(d). The distance-based neighbor searching algorithms lose the uniqueness of selecting nearest neighbors for each point when the key parameter is too large. On the other hand, the FC graph may bring too many connections for each point, which means the unique spatial features provided by the sets of neighbor points $\mathcal{N}(\cdot)$ are all the same for each point in this graph. Thus, the FC graph has the worst result (56.00\%) because it connects all the points in the graph. 

In conclusion, the Dstar ensures each point has the center point as its only neighbor, achieving near-optimal performance (94.27\%), focusing on the relative relationship between the center point and itself. Meanwhile, UStar loses the direction information, causing a drop in accuracy with (88.79\%). 
}

\subsubsection{Comparison with Non-graph-based Baseline Approaches} 

The MMPointGNN can get a high accuracy of overall 90.58\%, as shown in Table~\ref{tab:overall_accuracy}. It proves the spatial-temporal information of the human activity can be learned from the input mmWave radar point cloud by using GNNs~\cite{gong2021mmpoint}, but still struggles with complicated class 5, which only gets 50\% accuracy, as can be observed in Fig.~\ref{fig:class_accuracy}.

Voxelization involves the process of discretizing the continuous 3D space into a grid. Voxel can extract some spatial information from the voxelized point cloud but only gets 79.85\% test accuracy. It greatly increases the size of the data but has little contribution to spatial information extraction, which causes low accuracies at over half the classes (3, 4, 6, 10, 11). 

Meanwhile, RS and ZP have similar test accuracies with 80.65\% and 78.95\%, respectively. RS has low accuracies at classes 4, 6, and 11, while ZP has low accuracies at classes 3, 4, 5, and 6. RS and ZP use the PointNet model to individually process the points in each frame, but do not consider the relationships among points in each frame, making it hard to recognize these complicated classes. 

Because LR only uses the statistical features of each frame, it has the lowest accuracy, 51.37\%. It loses nearly all the spatial-temporal information in the point cloud.

\subsection{Inference Time via Raspberry Pi Implementation}
Traditional point cloud preprocessing and spatial feature extraction algorithms are mainly from vision-based point cloud areas, and they use powerful processors to accelerate the preprocessing speed. However, in real-world scenarios, mmWave radar is usually deployed with some resource-constraint devices. For example, a HAR system can be deployed in vehicles to detect driver behavior. In these circumstances, the computation ability of the processing unit is not as powerful as that of a modern PC with a GPU. Thus, we conducted an inference test with Raspberry Pi 4 because of its widespread use in the IoT applications~\cite{flamini2023prototype}. 

We first trained the DDGNN model on a PC and loaded the trained parameters on a Raspberry Pi 4 board for the inference test. The inference dataset was the same as the test dataset, but the batch size was set to 1 to simulate the real-world data processing scenario. We set up three metrics:  average Adjacency matrix generation time (Avg. Gen. Time) and average inference time (Avg. Infer. Time) for the model evaluations, and the inference accuracy (Inf. ACC). For the skeleton graph, we directly use the key point data from Kinect SDK, thus only considering the skeleton graph construction time. The results of all tested data types are summarized in Table~\ref{tab: inference test graph}.

\begin{table}[!t]
\caption{Implementation Performance on Raspberry Pi 4}
\begin{center}
\begin{tabular}{llcccc}
\hline
 & \textbf{Name}& \textbf{Avg. Gen.} & \textbf{Avg. Inf.}  & \textbf{Inf.}\\
 & & (ms) & (ms) &   \textbf{ACC} (\%) \\ 
\hline

\textbf{Ours} & DStar  & 25.80 & 137.60  & \textbf{93.99} \\
 & UStar & 25.87 & 143.58  & 88.24 \\
\hline

 &Skeleton & 50.20 & 139.94  & \textbf{97.78} \\
 &KNN & 76.84 & 163.63     & 90.71 \\
\textbf{Graph-based} & Radius  & 44.29 & 161.46  & 86.66  \\
\textbf{Baselines} &FC& 27.06 & 174.51  & 54.38 \\
 & Empty & 22.96 & 150.85 & 88.10  \\
\hline

 &MMPointGNN & 14.87 & 1486.00  & 90.59 \\
 &Voxel & 119.08 & 636.60   & 69.23 \\
\textbf{Non-graph-} & ZP & \textbf{14.59} & \textbf{26.19}   & 75.82 \\
\textbf{based Baselines} &RS & 165.21 & 56.63   & 80.65 \\
 &LR & 27.05 & 0.02   & 51.37 \\
\hline
\end{tabular}
\label{tab: inference test graph}
\end{center}
\end{table}

Overall, the inference accuracies reveal the similar performance as Table~\ref {tab:overall_accuracy}. DStar and UStar are similar in construction and inference time. However, compared to the skeleton graph, DStar provides higher accuracy with directed connections and achieves near-optimal performance of 93.99\%. Empty and UStar lose the essential relative information in connections and achieve 88.24\% and 88.10\% inference accuracy, respectively. 

KNN and radius graphs, due to their extra pairwise distance calculation, get high generation time and high inference time. Their inference accuracy results do not reach the Dstar performance, highlighting the effectiveness of the DStar representation. Empty and FC get generation times similar to DStar and Ustar since they do not calculate the pairwise distance. However, the FC graph has to consider all the connections in the point cloud, owning the highest inference time among all graph types. Overall, DDGNN with DStar reaches near-optimal performance compared to the Skelton graph and the other graph types.

As for MMPointGNN, we found that in the original paper, the author mentioned that due to the fully connected graph, MMPointGNN requires more computation resources during the training stage than inference. But in our inference test, it still has the highest inference time but the second-highest inference accuracy, as it requires searching for the most essential of connections across all the points across frames and requires a huge amount of time to reduce irrelevant information. Although the inference accuracy is guaranteed, the inference time is not suitable for practical applications.

Voxelization has the second-highest inference time and the second-highest construction time, as it requires building high-dimensional matrices. However, voxelized point clouds blur the spatial information in each frame because a single voxel contains multiple points if the points are close to each other, resulting in lower inference accuracy. ZP has the lowest construction and inference times among DL-based experiments, but its accuracy cannot be guaranteed. RS requires both upsampling and downsampling algorithms, leading to the highest construction time. Statistical-based LR is the fastest algorithm since it does not use DL-based models, but it suffers from low accuracy.

Overall, graph baselines with DDGNN have a similar construction time and a longer inference time than PointNet-based experiments (RS \& ZP) since GCNs require extra information aggregation procedures to capture relative relationships from variable-size inputs. However, the DDGNN model can achieve higher accuracy with proper graph representations, yet maintain acceptable complexity.

However, the current implementation of DDGNN still requires a long inference time (137.60 ms) compared to the fastest ZP baseline (26.19 ms). To further reduce computational overhead, quantization-aware training or metric-based model pruning \cite{chen2023efficient} can be used. Recently, some lightweight architectures, such as lightweight GNNs \cite{xie2024towards} and lightweight spatial-temporal networks \cite{zou2024stfnet}, have been proposed to optimize practical implementations on resource-limited platforms.

\sh{\subsection{Ablation Studies}
We now conduct the ablation studies to verify the necessity of the different modules in our DDGNN model.

\begin{figure}[!t]
\sh{
    \centering
    \subfloat[]{\includegraphics[height=0.94in]{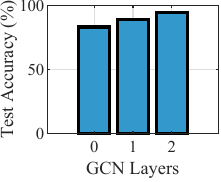}}
    \subfloat[]{\includegraphics[height=0.94in]{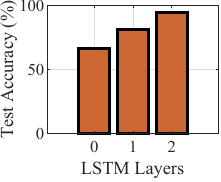}}
    \subfloat[]{\includegraphics[height=0.94in]{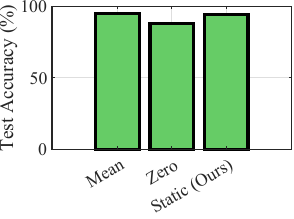}}
    \caption{\sh{(a) Different GCN layers; (b) Different LSTM layers; (c) Different choices of center point.}}
    \label{fig:ablation study}
    }
\end{figure}
\subsubsection{GCN Layers}
Fig.~\ref{fig:ablation study}(a) compares spatial models with 0, 1, and 2 layers, while the temporal decoder is fixed to two LSTM layers. The 0-layer case means we replace the GCN layer with a single FC layer. The network relies only on point-wise features and achieves $83.27\%$ accuracy. Adding a single GCN layer allows each point to observe its directed neighbors and improves the score to $89.02\%$. A two-layer stack further increases accuracy to $94.27\%$, because each point can aggregate richer spatial information from the first layer's representation learning.

\subsubsection{GCN Replacement}
\begin{table}[!t]
    \centering
    \sh{
    \caption{Overall Test Accuracies of GCN Replacements}
    \label{tab:variable_structure GCN}
    \begin{tabular}{lc}
    \hline
    \textbf{Model Configuration} & \textbf{Overall Accuracy (\%)} \\
    \hline
    DGCNN -LSTM & 81.83 \\
    PointMLP -LSTM  & 93.73 \\
    PointTransformer-LSTM & 86.80 \\
    PointNet++-LSTM &  87.84 \\
    \hline
    \textbf{DDGNN (Our Model)} & \textbf{94.27} \\
    \hline
    \end{tabular}
    }
\end{table}
    
We then replace the GCN encoder with alternative spatial feature extractors while maintaining the Bi-LSTM feature extractor. The overall test accuracy is shown in Table~\ref{tab:variable_structure GCN}. These models are used in recent mmWave-radar point cloud tasks \cite{cui2024milipoint, guo2023point, palipana2021pantomime}.

DGCNN \cite{cui2024milipoint} dynamically constructs the KNN graph based on each layer's representation of the points. It achieves $81.83\%$ accuracy. This drop suggests that the dynamic KNN graph construction in DGCNN may not be suitable for the sparse and variable-size point cloud. 

PointMLP \cite{cui2024milipoint} gets the overall accuracy at $93.73\%$. This model rethinks the point cloud analysis and provides a new residual point connect block structure for the spatial feature extraction. It also uses the KNN search to extract spatial features, but its residual point block performs better than our KNN baselines. 

PointTransformer \cite{guo2023point} reaches $86.80\%$. PointTransformer still relies on the KNN algorithm to build subsets for the point cloud, which encodes the relative relationship among different subsets in each frame. Such a structure may not be suitable for the variable-size point cloud.

PointNet++ \cite{palipana2021pantomime} is an improved version of the PointNet, and it has an improved accuracy (87.84\%) compared to the PointNet with ZP (75.82\%) and RS (80.65\%) baselines since it provides a radius-based KNN search algorithm for feature extraction in each frame. It still cannot reach the near-optimal performance under the variable-size point cloud.

\subsubsection{Bi-LSTM Layers}
Fig.~\ref{fig:ablation study}(b) compares the temporal model with 0, 1, and 2 layers, while the GCN depth is kept at two. A 0-layer Bi-LSTM means we replace the LSTM model with a single FC layer, which reduces accuracy to $66.54\%$, indicating that motion dynamics are crucial. A single LSTM layer recovers a large part of the gap and achieves $81.18\%$. Extending to two layers brings the score up to $94.27\%$. The extra layer helps the network capture longer temporal dependencies.

\subsubsection{Bi-LSTM Replacement} \label{sec: Bi-LSTM Replacements}
\begin{table}[!t]
    \centering
    \sh{
    \caption{Overall Test Accuracies of Bi-LSTM Replacements}
    \label{tab:variable_structure LSTM}
    \begin{tabular}{lc}
    \hline
    \textbf{Model Configuration} & \textbf{Overall Accuracy (\%)} \\
    \hline
    GCN-Bi-GRU & 93.33 \\
    GCN-Transformer & 80.52 \\
    
    \hline
    \textbf{DDGNN (Our Model)} & \textbf{94.27} \\
    \hline
    \end{tabular}
    }
    \end{table}
We then replace the Bi-LSTM module with Bi-GRU and Transformer while keeping the GCN encoder. The overall test accuracy is given in Table~\ref{tab:variable_structure LSTM}. We choose these two temporal models for comparison because they are also widely used in mmWave radar HAR tasks \cite{wang2021m, guo2023point}. For consistency considerations, Bi-GRU has two layers and the Transformer has two feedforward layers.

Bi-GRU achieves $93.33\%$ accuracy, which is very close to our method ($94.27\%$). However, the Transformer performs at $80.52\%$, which does not perform as well as our method and Bi-GRU in this task. The results show that the choice of temporal feature extractor is crucial for the overall performance.

\subsubsection{Choices of Center Point}
We tested three choices for the center point of the directed star graph in each frame:
\begin{enumerate}
    \item \emph{Mean}: an additional spatial-mean point of the point cloud is used as the center point.
    \item \emph{Zero}: an all-zero point is used as the center point.
    \item \emph{Static (Ours)}: a fixed non-zero point (e.g., (0,1,0)) is used as the center point.
\end{enumerate}
The results are shown in Fig.~\ref{fig:ablation study}(c). The overall accuracy of the \emph{Mean} method reaches $94.77\%$ and our method with \textit{static} center point achieves $94.27\%$. The \emph{Zero} method is $88.10\%$. \emph{Mean} provides a statistical average center for each frame, while our method provides a constant and static center for each frame. Both methods provide a meaningful relative relationship in each frame compared to the \emph{Zero} method. These results show the importance of a meaningful and stable center point in the star graph construction. 
}

\sh{\subsection{Comparisons to Recent Methods}\label{sec: results on SOTA}
We also compare the proposed DDGNN models with the three recent methods. The results are shown in Table~\ref{tab:variable_structure}.

\begin{table}[!t]
\centering
\sh{
\caption{Overall Test Accuracy of Recent Methods}
\label{tab:variable_structure}
\begin{tabular}{lc}
\hline
\textbf{Model Configuration} & \textbf{Overall Accuracy (\%)} \\
\hline
ST-PCT \cite{kang2023st} & 82.47 \\
mmPoint-Attention \cite{shi2025mmpoint} & 85.75\\
Tesla-Rapture \cite{salami2022tesla} & 87.77 \\
\hline
\textbf{DDGNN (Our Model)} & \textbf{94.27} \\
\hline
\end{tabular}
}

\end{table}
ST-PCT \cite{kang2023st} refers to the Spatial–Temporal Point Cloud Transformer. The algorithm uses both random sampling and frame aggregation for preprocessing. The spatial-temporal Attention Encoder with KNN graph and FPS sampling provides a comprehensive feature extraction structure. It achieves 82.47\% overall test accuracy, showing that this attention module with KNN graph may not be suitable for variable-size point clouds. 

mmPoint-Attention \cite{shi2025mmpoint} introduces a unified attention model for mmWave radar point-cloud-based HAR. It requires increasing the point cloud intensity by using a frame aggregator and zero-padding during preprocessing. Then, a CNN is used as a spatial feature extractor, and a transformer with efficient channel attention and sequential-weighted attention is employed as both a spatial-temporal feature encoder. It achieves an accuracy of 85.75\%, which is lower than our DStar results (94.27\%).

Tesla-Rapture \cite{salami2022tesla} introduces a frame-aggregator and resampling algorithms for the preprocessor. Then, a specialized Tesla-Net with attention modules and a temporal-KNN graph is proposed for the spatial-temporal feature extraction. It gets 87.77\% accuracy on our test dataset.

In conclusion, our DDGNN with DStar performs better than these attention-enhanced methods without using preprocessing algorithms like frame-aggregator or resampling algorithms. 
}

\section{Complexity Analysis and Discussions}\label{sec: analysis}
In this section, we present a complexity analysis for the graph construction, GCN module, and the Bi-LSTM module with Big-$O$ time (operation count) and space (memory requirements) complexities \cite{wu2020comprehensive}. 

Then, we will discuss the generalizations of the mmWave-radar HAR to complex scenarios.

\subsection{Complexity Analysis of Graph Constructions}
The complexity of GCN directly relies on the number of edges in the graph. Thus, before we analyze the GCN operation, it is necessary to analyze the complexity in different graphs.
Table~\ref{tab:graph_time_space} summarizes the time and space complexities for the graph-based baselines. The time complexity reflects the cost of graph construction per frame. Meanwhile, the space complexity reflects the memory needed to store the adjacency matrix and per-point features required during computation.
\begin{table}[!t]
\centering

\caption{Time and Space Complexity of Different Graphs}
\label{tab:graph_time_space}
\begin{tabular}{lcc}
\hline
\textbf{Graph Type} & \textbf{Time Complexity} & \textbf{Space Complexity} \\
\hline
Star Graph (ours)    
  & \(O(N_{t_n})\)  
  & \(O(N_{t_n})\) \\

  Skeleton Graph (vision-based)        
  & \(O(N_{skeleton})\)  
  & \(O(\mathcal{E}_{skeleton})\) \\

KNN Graph              
  & \(O(N_{t_n}^2)\)  
  & \(O(\mathcal{E}_{t_n})\) \\

Radius Graph           
  & \(O(N_{t_n}^2)\)  
  & \(O(\mathcal{E}_{t_n})\) \\

FC Graph  
  & \(O(N_{t_n}^2)\)  
  & \(O(N_{t_n}^2)\) \\
Empty Graph            
  & \(O(N_{t_n})\)  
  & \(O(N_{t_n})\) \\
\hline
\end{tabular}
\end{table}

For the time complexity, the empty graph and star graph have only linear complexity with input $N_{t_n}$ since they only calculate the number of points in each frame once. Meanwhile, the KNN graph and Radius Graph have higher quadratic complexity of $O(N_{t_n}^2)$ since they need to calculate the pair-wise distance between all point-pairs in the frame. FC graph also has the quadratic complexity since it needs to connect all the point pairs in the same frame. Skeleton graph only calculates the number of skeleton points in one frame, so it has linear complexity. 

For the space complexity, the star graph stores $N_{t_n}$ connections in each frame. An empty graph has no connections but it needs to store at least $N_{t_n}$ points for each frame. KNN and Radius graphs are determined by $\mathcal{E}_{t_n}$, which depends on the distance-based neighbor search algorithms. FC graph always has quadratic complexity since all the point pairs are connected, doubling the storage. The skeleton graph's complexity $\mathcal{E}_{skeleton}$ which are determined by the human skeletal structure.

\subsection{Complexity Analysis of DDGNN}
\subsubsection{GCN Operation}

For a GraphConv layer shown in (\ref{eq: graphconv 2}), the time complexity are determined by three parameters: $\mathcal{E}_{t_n}, F_{\mathrm{in}}^{(l)}, F_{\mathrm{out}}^{(l)}$ \cite{kipf2016semi, paszke2019pytorch}
, represented as $ O\big(\mathcal{E}_{t_n}\,F_{\mathrm{in}}^{(l)}\,F_{\mathrm{out}}^{(l)}\big)$. The space complexity is $O\big(\mathcal{E}_{t_n}\big)$ since it stores the aggregated features from each connection. 

The time and space complexity of the GraphConv module scales linearly with $\mathcal{E}_{t_n}$. When using the star graph, both time and memory scales linearly with $N_{t_n}$, which makes per-frame computation predictable and memory-friendly even as the number of radar points grows \cite{kipf2016semi}.

\subsubsection{Bi-LSTM Temporal Modeling}
For a single LSTM cell at one time step, the computation involves four gates (input, forget, output, and candidate), each requiring matrix multiplications. Thus, the time complexity per time step has a quadratic complexity of $O\big(F_{\mathrm{LSTM}}^2\big)$. This remains the same for the forward and backward directions, thus the total time complexity per time step of Bi-LSTM is still $O\big( F_{\mathrm{LSTM}}^2\big)$.

For temporal modeling of a complete sequence of length $N$, we must process all time steps. For each layer, the total time complexity becomes $O\big( N  F_{\mathrm{LSTM}}^2\big)$. 
For the space complexity, temporal modeling requires storing hidden states across the entire sequence to maintain temporal dependencies $O(N)$.

The Bi-LSTM's computational cost scales \emph{linearly} with sequence length $N$, making it suitable for moderate-length radar sequences. In contrast, Transformer-based methods scale \emph{quadratically} with sequence length (e.g., $O(N^2)$ for self-attention) \cite{yan2022mm}, which becomes prohibitive for long point cloud sequences. For our dataset with $T=50$ and $H_t=32$, Bi-LSTM provides an optimal temporal modeling choice while maintaining competitive performance as demonstrated in Section~\ref{sec: results}.

\sh{\subsection{Discussions}\label{sec: discussion}}
In this section, we discuss some generalizations to complex real-world scenarios.

\subsubsection{Complex Activity Scenarios}
The current dataset only contains 13 activities. Although it includes some difficult activities like 11 and 12, it does not contain more complex activities such as bending or sitting. Collecting massive training data requires huge human effort. This can be solved by applying a synthetic dataset \cite{chen2023rf} to generate more complex activities. Also, a self-supervised pretraining strategy \cite{crossharhong2024} can combine multiple datasets with different activities.
\sh{
\subsubsection{Multi-target Activity Recognition Scenarios}
As mmWave radar can sense multiple moving targets simultaneously, some recent works with mmWave radar \cite{alam2021palmar, wang2025multi} separate the multi-person point cloud into multiple single-person subsets and then individually process each subgroup as a single-person HAR.
}
\subsubsection{Human Interaction Scenarios}
Human interaction refers to two scenarios: Human-to-Computer and Human-to-Human. Human-to-Computer Interactions always contain continuous activities, which may require real-time segmentation to extract each activity. ZuMa, introduced in \cite{liu2024real}, provides a real-time segmentation strategy for activities with varying durations.  Human-to-human interaction scenario shares a similar pipeline to multi-target activity recognition scenarios \cite{zhang2025mmhiu} that requires separately processing each human target. 

\subsubsection{Occlusion Scenarios}
mmWave radar can operate under occlusion scenarios, such as when penetrating non-metallic objects \cite{wang2025open}. In these scenarios, the mmWave radar point cloud becomes more sparse because materials and obstacles weaken the reflected signal strength. A novel data augmentation strategy for mmWave radar sparse point clouds was designed in \cite{wang2025open} to increase the dynamic information related to human motion under occlusion scenarios.

\subsubsection{Dynamic Environment Scenarios}
Recently, some works have been exploring the application of mmWave radar in dynamic scenarios. HAR in dynamic scenarios is non-trivial since human movement may cause different patterns of human activities compared to the static human body. Wang \textit{et al} in \cite{wang2023real} provide a simultaneous human localization and activity recognition algorithm with a well-designed multi-feature fusion network in dynamic scenarios.

\sh{
\subsubsection{Outdoor Scenarios}
Implementing mmWave radar-based HAR in outdoor scenarios has been explored in 
\cite{luo2019human, deng2023midas}. Authors in \cite{luo2019human} consider that coarse localization is essential in sparse urban areas for HAR. Experiments in \cite{deng2023midas} show that generating synthetic outdoor mmWave radar points from videos can perform well in real-world outdoor test scenarios.
}

\section{Related Works} \label{sec: related works}
In this section, we introduce related works on skeleton-based HAR tasks and discuss their extended applications in mmWave radar point cloud-based works.
\subsection{Skeleton Graph in Vision-based Tasks}
In Section \ref{sec: problem_state}, we have introduced one of the most classical solutions in skeleton-based HAR: the skeleton graph and ST-GCN \cite{yan2018spatial}. Recently, many advanced solutions for skeleton-based HAR have been proposed. Chen \textit{et al} in \cite{chen2021channel} design a new channel-wise Topology Refinement Graph Convolution to dynamically learn different joint features in different channels. \cite{tu2022joint} fuses the motion features of the joints and the bones, and designs a temporal prediction head for self-supervised skeleton feature mining. Zhang \textit{et al} in \cite{zhang2022zoom} explore more complicated scenarios, such as group activity recognition tasks, and create a Zoom-in \& out Transformer to explore high-level semantic information from multiple skeleton graphs. Pang \textit{et al} in \cite{pang2023self} introduce a self-adaptive GCN (SAGCN) with a global attention network to dynamically search for connections between skeleton joints in a single frame and capture the comprehensive temporal features from the sequence.  
\subsection{Skeleton Graph in mmWave Radar Tasks}
As we previously discussed, directly generating a skeleton graph from a raw mmWave radar point cloud is not possible. Current works utilize vision-based skeleton points as a reference to train an encoder that learns a mapping function from the raw mmWave radar point cloud to the skeleton points for human pose estimation purposes \cite{wang2023human, sengupta2022mmpose}. Then, HAR can be considered a downstream task of the pose estimation task.

\section{Conclusion} \label{sec: conclusion}
In this paper, we proposed a novel star graph with a DDGNN model to explore the spatial-temporal representation in the HAR system based on the mmWave radar point cloud. We explored the relative relationship between the manually added static center point and dynamic radar points in the star graph across frames. We also provided a comprehensive analysis of the representation learning of two different star graphs. We designed a DDGNN model to extract the spatial-temporal features from the variable-size input graphs. Experiment results revealed that star graph representation, specifically the directed star graph, can achieve a 94.27\% test accuracy on the real-world collected HAR dataset. It outperformed the other graph-based and the non-graph-based baseline approaches. It also reached the near-optimal performance as the vision-based skeleton graph. Moreover, we implemented our star graph-based algorithms and baseline approaches on Raspberry Pi 4. The inference test showed our star graph-based systems have affordable complexity, which is very suitable for resource-constrained embedded platforms. Ablation studies and comparisons with recent radar-specific methods enhance the effectiveness of our system. We then provided a series of complexity analyses and discussions of generalizations for our systems.


\bibliographystyle{IEEEtran}
\bibliography{IEEEabrv,reference}

@STRING{IEEE_J_IV         = "{IEEE} Trans. Intell. Veh."}

@STRING{IEEE_J_VT         = "{IEEE} Trans. Veh. Technol."}

@STRING{IEEE_J_SYST       = "{IEEE} Syst. J."}

@STRING{IEEE_J_CASVT      = "{IEEE} Trans. Circuits Syst. Video Technol."}

@STRING{IEEE_J_IOT        = "{IEEE} Internet Things J."}

@STRING{IEEE_J_MC         = "{IEEE} Trans. Mobile Comput."}

@STRING{IEEE_J_NNLS       = "{IEEE} Trans. Neural Netw. Learn. Syst."}

@STRING{IEEE_J_MM         = "{IEEE} Trans. Multimedia"}

@STRING{IEEE_J_PAMI       = "{IEEE} Trans. Pattern Anal. Mach. Intell."}

@STRING{IEEE_J_GRS        = "{IEEE} Trans. Geosci. Remote Sens."}

@STRING{IEEE_J_IA         = "{IEEE} Trans. Ind. Appl."}

@STRING{IEEE_J_SENSOR     = "{IEEE} Sensors J."}

@STRING{IEEE_O_CSTO       = "{IEEE} Commun. Surveys Tuts."}

@STRING{IEEE_M_MM         = "{IEEE} Multimedia"}

@inproceedings{chen203universal,
  title={Universal Targeted Adversarial Attacks Against mm{W}ave-based Human Activity Recognition},
  author={Xie, Yucheng and Guo, Xiaonan and Wang, Yan and Cheng, Jerry and Chen, Yingying},
  booktitle={Proc. IEEE INFOCOM},
  year={2023}
}

@inproceedings{chen2021channel,
  title={Channel-wise topology refinement graph convolution for skeleton-based action recognition},
  author={Chen, Yuxin and Zhang, Ziqi and Yuan, Chunfeng and Li, Bing and Deng, Ying and Hu, Weiming},
  booktitle={Proc. IEEE/CVF ICCV},
  pages={13359--13368},
  year={2021}
}

@article{luo2019human,
  title={Human activity detection and coarse localization outdoors using micro-Doppler signatures},
  author={Luo, Fei and Poslad, Stefan and Bodanese, Eliane},
  journal=IEEE_J_SENSOR,
  volume={19},
  number={18},
  pages={8079--8094},
  year={2019},
}

@article{tu2022joint,
  title={Joint-bone fusion graph convolutional network for semi-supervised skeleton action recognition},
  author={Tu, Zhigang and Zhang, Jiaxu and Li, Hongyan and Chen, Yujin and Yuan, Junsong},
  journal=IEEE_J_MM,
  volume={25},
  pages={1819--1831},
  year={2022},
}

@article{pang2023self,
  title={Self-adaptive graph with nonlocal attention network for skeleton-based action recognition},
  author={Pang, Chen and Gao, Xingyu and Chen, Zhenyu and Lyu, Lei},
  journal=IEEE_J_NNLS,
  year={2023},
}

@article{zhang2022zoom,
  title={Zoom transformer for skeleton-based group activity recognition},
  author={Zhang, Jiaxu and Jia, Yifan and Xie, Wei and Tu, Zhigang},
  journal=IEEE_J_CASVT,
  volume={32},
  number={12},
  pages={8646--8659},
  year={2022},
}

@article{venon2022millimeter,
  title={Millimeter wave {FMCW} radars for perception, recognition and localization in automotive applications: A survey},
  author={Venon, Arthur and Dupuis, Yohan and Vasseur, Pascal and Merriaux, Pierre},
  journal=IEEE_J_IV,
  volume={7},
  number={3},
  pages={533--555},
  year={2022},
}

@article{wang2023real,
  title={Real-time through-wall multihuman localization and behavior recognition based on {MIMO} radar},
  author={Wang, Changlong and Zhu, Dongsheng and Sun, Lijuan and Han, Chong and Guo, Jian},
  journal=IEEE_J_GRS,
  volume={61},
  pages={1--12},
  year={2023},
}

@article{wang2024multi,
  title={Multi-Modal Fusion Sensing: A Comprehensive Review of {M}illimeter-Wave Radar and Its Integration With Other Modalities},
  author={Wang, Shuai and Mei, Luoyu and Liu, Ruofeng and Jiang, Wenchao and Yin, Zhimeng and Deng, Xianjun and He, Tian},
  journal=IEEE_O_CSTO,
  volume={},
  number={},
  pages={1-1},
  year={2024},
}

@inproceedings{wang2021m,
  title={m-activity: Accurate and real-time human activity recognition via millimeter wave radar},
  author={Wang, Yuheng and Liu, Haipeng and Cui, Kening and Zhou, Anfu and Li, Wensheng and Ma, Huadong},
  booktitle={Proc. IEEE ICASSP},
  pages={8298--8302},
  year={2021},
}

@inproceedings{gill2011system,
  title={A system for change detection and human recognition in voxel space using the {Microsoft Kinect} sensor},
  author={Gill, Tyler and Keller, James M and Anderson, Derek T and Luke, RH},
  booktitle={Proc. IEEE AIPR},
  pages={1--8},
  year={2011},

}

@article{zhang2023survey,
  title={A survey of mm{W}ave-based human sensing: Technology, platforms and applications},
  author={Zhang, Jia and Xi, Rui and He, Yuan and Sun, Yimiao and Guo, Xiuzhen and Wang, Weiguo and Na, Xin and Liu, Yunhao and Shi, Zhenguo and Gu, Tao},
  journal=IEEE_O_CSTO,
  volume={25},
  number={4},
  pages={2052-2087},
  year={2023},

}

@article{an2021mars,
  title={Mars: mm{W}ave-based assistive rehabilitation system for smart healthcare},
  author={An, Sizhe and Ogras, Umit Y},
  journal={ACM Trans. Embed. Comput. Syst.},
  volume={20},
  number={5s},
  pages={1--22},
  year={2021},

}

@article{yu2022noninvasive,
  title={Noninvasive human activity recognition using millimeter-wave radar},
  author={Yu, Chengxi and Xu, Zhezhuang and Yan, Kun and Chien, Ying-Ren and Fang, Shih-Hau and Wu, Hsiao-Chun},
  journal=IEEE_J_SYST,
  volume={16},
  number={2},
  pages={3036--3047},
  year={2022},
}

@article{sengupta2022mmpose,
  title={mmpose-{NLP}: A natural language processing approach to precise skeletal pose estimation using mm{W}ave radars},
  author={Sengupta, Arindam and Cao, Siyang},
  journal=IEEE_J_NNLS,
  volume={34},
  number={11},
  pages={8418--8429},
  year={2022},
}

@inproceedings{paszke2019pytorch,
  title={Pytorch: An imperative style, high-performance deep learning library},
  author={Paszke, Adam and Gross, Sam and Massa, Francisco and Lerer, Adam and Bradbury, James and Chanan, Gregory and Killeen, Trevor and Lin, Zeming and Gimelshein, Natalia and Antiga, Luca and others},
  booktitle={Proc. Neur{IPS}},
  volume={32},
  year={2019}
}

@article{shastri2022review,
  title={A review of millimeter wave device-based localization and device-free sensing technologies and applications},
  author={Shastri, Anish and Valecha, Neharika and Bashirov, Enver and Tataria, Harsh and Lentmaier, Michael and Tufvesson, Fredrik and Rossi, Michele and Casari, Paolo},
  journal=IEEE_O_CSTO,
  volume={25},
  number={4},
  pages={2052-2087},
  year={2022},
}

@article{deng2023midas,
  title={Midas: Generating mm{W}ave radar data from videos for training pervasive and privacy-preserving human sensing tasks},
  author={Deng, Kaikai and Zhao, Dong and Han, Qiaoyue and Zhang, Zihan and Wang, Shuyue and Zhou, Anfu and Ma, Huadong},
  journal={Proc. ACM IMWUT},
  volume={7},
  number={1},
  pages={1--26},
  year={2023},

}

@inproceedings{chen2023rf,
  title={{RF} genesis: Zero-shot generalization of mm{W}ave sensing through simulation-based data synthesis and generative diffusion models},
  author={Chen, Xingyu and Zhang, Xinyu},
  booktitle={Proc. ACM SenSys 2023},
  pages={28--42},
  year={2023}
}

@article{shi2025mmpoint,
  title={mmPoint-{A}ttention: A Unified Attention Framework for Human Pose and Activity Recognition From mm{W}ave Radar Point Clouds},
  author={Shi, Xintong and Bouazizi, Mondher and Ohtsuki, Tomoaki},
  journal=IEEE_J_SENSOR,
  year={2025},
}

@article{wang2025multi,
  title={Multi-human Activity Recognition based on Sequential {4D} Point Clouds using Frequency-modulated Continuous Wave Radar},
  author={Wang, Yong and Kong, Weishuo and Zhou, Mu and Nie, Wei and He, Wei and Zhang, Qian and Pang, Yu},
  journal=IEEE_J_VT,
  year={2025},
}

@article{liu2024real,
  title={Real-time continuous activity recognition with a commercial mm{W}ave radar},
  author={Liu, Yunhao and Zhang, Jia and Chen, Yande and Wang, Weiguo and Yang, Songzhou and Na, Xin and Sun, Yimiao and He, Yuan},
  journal=IEEE_J_MC,
  year={2024},
}

@inproceedings{wang2023human,
  title={Human Parsing with Joint Learning for Dynamic mm{W}ave Radar Point Cloud},
  author={Wang, Shuai and Cao, Dongjiang and Liu, Ruofeng and Jiang, Wenchao and Yao, Tianshun and Lu, Chris Xiaoxuan},
  booktitle={Proc. ACM IMWUT},
  volume={7},
  number={1},
  pages={1--22},
  year={2023},
}

@article{ti,
  title={Introduction to mm{W}ave sensing: {FMCW} radars},
  author={Rao, Sandeep},
  journal={Texas Instruments {(TI)} mm{W}ave Training Series},
  pages={1--11},
  year={2017},

}

@inproceedings{singh2019radhar,
  title={Radhar: Human activity recognition from point clouds generated through a millimeter-wave radar},
  author={Singh, Akash Deep and Sandha, Sandeep Singh and Garcia, Luis and Srivastava, Mani},
  booktitle={Proc. ACM Workshop on mm{N}ets},
  pages={51--56},
  year={2019}
}

@inproceedings{liu2019point,
  title={Point-voxel {CNN} for efficient {3D} deep learning},
  author={Liu, Zhijian and Tang, Haotian and Lin, Yujun and Han, Song},
  booktitle={Proc. Neur{IPS}},
  volume={32},
  year={2019}
}

@inproceedings{guo2023point,
  title={Point transformer-based human activity recognition using high-dimensional radar point clouds},
  author={Guo, Zhongyuan and Guendel, Ronny G and Yarovoy, Alexander and Fioranelli, Francesco},
  booktitle={Proc. IEEE RadarConf},
  pages={1--6},
  year={2023},
}

@inproceedings{xie2024towards,
  title={Towards lightweight graph neural network search with curriculum graph sparsification},
  author={Xie, Beini and Chang, Heng and Zhang, Ziwei and Zhang, Zeyang and Wu, Simin and Wang, Xin and Meng, Yuan and Zhu, Wenwu},
  booktitle={Proc. ACM SIGKDD},
  pages={3563--3573},
  year={2024}
}

@inproceedings{yan2022mm,
  title={Mm-hat: {T}ransformer for millimeter-wave sensing based human activity recognition},
  author={Yan, Jie and Zeng, Xianlin and Zhou, Anfu and Ma, Huadong},
  booktitle={Proc. IEEE GLOBECOM},
  pages={547--553},
  year={2022},
}

@article{wang2025open,
  title={Open-set occluded person identification with mm{W}ave radar},
  author={Wang, Tao and Zhao, Yang and Chang, Ming-Ching and Liu, Jie},
  journal=IEEE_J_MC,
  year={2025},
}

@article{zou2024stfnet,
  title={{STFN}et: Enhanced and lightweight spatiotemporal fusion network for wearable human activity recognition},
  author={Zou, Hailin and Chen, Zijie and Zhang, Chenyang and Yuan, Anran and Wang, Binbin and Wang, Lei and Li, Jianqing and Pan, Yuanyuan},
  journal=IEEE_J_SENSOR ,
  volume={24},
  number={8},
  pages={13686--13698},
  year={2024},

}

@inproceedings{chen2023efficient,
  title={Efficient Multi-channel Automotive Radar Interference Mitigation Using Pruned and Quantized Neural Networks},
  author={Chen, Shengyi and Klemp, Marvin and Taghia, Jalal and Martin, Rainer},
  booktitle={Proc. IEEE {RADAR}},
  pages={1--6},
  year={2023},
}

@inproceedings{shao2020finegym,
  title={Finegym: A hierarchical video dataset for fine-grained action understanding},
  author={Shao, Dian and Zhao, Yue and Dai, Bo and Lin, Dahua},
  booktitle={Proc. IEEE/CVF CFPR},
  pages={2616--2625},
  year={2020}
}

@article{wang2019dynamic,
  title={Dynamic graph {CNN} for learning on point clouds},
  author={Wang, Yue and Sun, Yongbin and Liu, Ziwei and Sarma, Sanjay E and Bronstein, Michael M and Solomon, Justin M},
  journal={ACM Trans. Graph. },
  pages={1--12},
  year={2019},

}

@article{gu2021survey,
  title={A survey on deep learning for human activity recognition},
  author={Gu, Fuqiang and Chung, Mu-Huan and Chignell, Mark and Valaee, Shahrokh and Zhou, Baoding and Liu, Xue},
  journal={ACM Comput. Surv.},
  volume={54},
  number={8},
  pages={1--34},
  year={2021},
}

@inproceedings{liu2019meteornet,
  title={Meteornet: Deep learning on dynamic 3d point cloud sequences},
  author={Liu, Xingyu and Yan, Mengyuan and Bohg, Jeannette},
  booktitle={Proc. IEEE/CVF ICCV},
  pages={9246--9255},
  year={2019}
}

@article{zhang2012microsoft,
  title={Microsoft {K}inect sensor and its effect},
  author={Zhang, Zhengyou},
  journal=IEEE_M_MM,
  volume={19},
  number={2},
  pages={4--10},
  year={2012},

}

@inproceedings{palipana2021pantomime,
  title={Pantomime: Mid-air gesture recognition with sparse millimeter-wave radar point clouds},
  author={Palipana, Sameera and Salami, Dariush and Leiva, Luis A and Sigg, Stephan},
  booktitle={Proc. ACM IMWUT},
  volume={5},
  number={1},
  pages={1--27},
  year={2021},

}

@article{salami2022tesla,
  title={Tesla-rapture: A lightweight gesture recognition system from mm{W}ave radar sparse point clouds},
  author={Salami, Dariush and Hasibi, Ramin and Palipana, Sameera and Popovski, Petar and Michoel, Tom and Sigg, Stephan},
  journal=IEEE_J_MC,
  volume={22},
  number={8},
  pages={4946--4960},
  year={2022},

}

@article{liu2019wireless,
  title={Wireless sensing for human activity: A survey},
  author={Liu, Jian and Liu, Hongbo and Chen, Yingying and Wang, Yan and Wang, Chen},
  journal=IEEE_O_CSTO,
  volume={22},
  number={3},
  pages={1629--1645},
  year={2019},

}

@inproceedings{min2020efficient,
  title={An efficient {PointLSTM} for point clouds based gesture recognition},
  author={Min, Yuecong and Zhang, Yanxiao and Chai, Xiujuan and Chen, Xilin},
  booktitle={Proc. IEEE/CVF CVPR},
  pages={5761--5770},
  year={2020}
}

@inproceedings{qi2017pointnet,
  title={Pointnet: Deep learning on point sets for 3d classification and segmentation},
  author={Qi, Charles R and Su, Hao and Mo, Kaichun and Guibas, Leonidas J},
  booktitle={Proc. IEEE/CVF CVPR},
  pages={652--660},
  year={2017}
}

@inproceedings{gong2021mmpoint,
  title={Mmpoint-gnn: Graph neural network with dynamic edges for human activity recognition through a millimeter-wave radar},
  author={Gong, Peixian and Wang, Chunyu and Zhang, Lihua},
  booktitle={Proc. IEEE IJCNN},
  pages={1--7},
  year={2021},

}

@article{guo2020deep,
  title={Deep learning for {3D} point clouds: A survey},
  author={Guo, Yulan and Wang, Hanyun and Hu, Qingyong and Liu, Hao and Liu, Li and Bennamoun, Mohammed},
  journal=IEEE_J_PAMI,
  volume={43},
  number={12},
  pages={4338--4364},
  year={2020},

}

@inproceedings{cui2024milipoint,
  title={Mili{P}oint: A Point Cloud Dataset for mm{W}ave Radar},
  author={Cui, Han and Zhong, Shu and Wu, Jiacheng and Shen, Zichao and Dahnoun, Naim and Zhao, Yiren},
  booktitle={Proc. Neur{IPS}},
  volume={36},
  year={2024}
}

@inproceedings{zhou2018voxelnet,
  title={Voxelnet: End-to-end learning for point cloud based 3d object detection},
  author={Zhou, Yin and Tuzel, Oncel},
  booktitle={Proc. IEEE/CVF CVPR},
  pages={4490--4499},
  year={2018}
}

@inproceedings{shi2019skeleton,
  title={Skeleton-based action recognition with directed graph neural networks},
  author={Shi, Lei and Zhang, Yifan and Cheng, Jian and Lu, Hanqing},
  booktitle={Proc. IEEE/CVF CVPR},
  pages={7912--7921},
  year={2019}
}

@article{presti20163d,
  title={{3D} skeleton-based human action classification: A survey},
  author={Presti, Liliana Lo and La Cascia, Marco},
  journal={Pattern Recognition},
  volume={53},
  pages={130--147},
  year={2016},

}

@inproceedings{yan2018spatial,
  title={Spatial temporal graph convolutional networks for skeleton-based action recognition},
  author={Yan, Sijie and Xiong, Yuanjun and Lin, Dahua},
  booktitle={Proc. AAAI},
  volume={32},
  number={1},
  year={2018}
}

@inproceedings{cheng2020skeleton,
  title={Skeleton-based action recognition with shift graph convolutional network},
  author={Cheng, Ke and Zhang, Yifan and He, Xiangyu and Chen, Weihan and Cheng, Jian and Lu, Hanqing},
  booktitle={Proc. IEEE/CVF CVPR},
  pages={183--192},
  year={2020}
}

@inproceedings{wu2022graph,
  title={Graph neural networks: foundation, frontiers and applications},
  author={Wu, Lingfei and Cui, Peng and Pei, Jian and Zhao, Liang and Guo, Xiaojie},
  booktitle={Proc. ACM SIGKDD},
  pages={4840--4841},
  year={2022}
}

@article{chen2021deep,
  title={Deep learning for sensor-based human activity recognition: Overview, challenges, and opportunities},
  author={Chen, Kaixuan and Zhang, Dalin and Yao, Lina and Guo, Bin and Yu, Zhiwen and Liu, Yunhao},
  journal={ACM Comput. Surv.},
  volume={54},
  number={4},
  pages={1--40},
  year={2021},

}

@article{kipf2016semi,
  title={Semi-supervised classification with graph convolutional networks},
  author={Kipf, Thomas N and Welling, Max},
  journal={arXiv preprint arXiv:1609.02907},
  year={2016}
}

@article{sak2014long,
  title={Long short-term memory based recurrent neural network architectures for large vocabulary speech recognition},
  author={Sak, Ha{\c{s}}im and Senior, Andrew and Beaufays, Fran{\c{c}}oise},
  journal={arXiv preprint arXiv:1402.1128},
  year={2014}
}

@article{shrestha2020continuous,
  title={Continuous human activity classification from {FMCW} radar with {B}i-{LSTM} networks},
  author={Shrestha, Aman and Li, Haobo and Le Kernec, Julien and Fioranelli, Francesco},
  journal=IEEE_J_SENSOR,
  volume={20},
  number={22},
  pages={13607--13619},
  year={2020},
}

@article{flamini2023prototype,
  title={A prototype of low-cost home automation system for energy savings and living comfort},
  author={Flamini, Alessandro and Ciurluini, Luca and Loggia, Riccardo and Massaccesi, Andrea and Moscatiello, Cristina and Martirano, Luigi},
  journal=IEEE_J_IA,
  volume={59},
  number={4},
  pages={4931--4941},
  year={2023},

}

@book{webb2012beginning,
  title={Beginning {K}inect programming with the {M}icrosoft {K}inect {SDK}},
  publisher={Apress},
  author={Webb, Jarrett and Ashley, James},
  year={2012},
}

@article{kang2023st,
  title={{ST-PCT}: Spatial-Temporal Point Cloud Transformer for Sensing Activity Based on mm{W}ave},
  author={Kang, Liyu and Li, Zan and Zhao, Xiaohui and Zhao, Zhongliang and Braun, Torsten},
  journal=IEEE_J_IOT,
  volume={11},
  number={6},
  pages={10979-10991},
  year={2023},

}

@inproceedings{franceschi2019learning,
  title={Learning discrete structures for graph neural networks},
  author={Franceschi, Luca and Niepert, Mathias and Pontil, Massimiliano and He, Xiao},
  booktitle={Proc. ICML},
  pages={1972--1982},
  year={2019},

}

@inproceedings{qi2017pointnet++,
  title={Pointnet++: Deep hierarchical feature learning on point sets in a metric space},
  author={Qi, Charles Ruizhongtai and Yi, Li and Su, Hao and Guibas, Leonidas J},
  booktitle={Proc. AAAI},
  volume={30},
  year={2017}
}

@article{ahmad2021graph,
  title={Graph convolutional neural network for human action recognition: A comprehensive survey},
  author={Ahmad, Tasweer and Jin, Lianwen and Zhang, Xin and Lai, Songxuan and Tang, Guozhi and Lin, Luojun},
  journal={IEEE Trans. Artif. Intell.},
  volume={2},
  number={2},
  pages={128--145},
  year={2021},
}

@inproceedings{morris2019weisfeiler,
  title={Weisfeiler and {L}eman go neural: Higher-order graph neural networks},
  author={Morris, Christopher and Ritzert, Martin and Fey, Matthias and Hamilton, William L and Lenssen, Jan Eric and Rattan, Gaurav and Grohe, Martin},
  booktitle={Proc. AAAI},
  volume={33},
  number={01},
  pages={4602--4609},
  year={2019}
}

@article{wu2020comprehensive,
  title={A comprehensive survey on graph neural networks},
  author={Wu, Zonghan and Pan, Shirui and Chen, Fengwen and Long, Guodong and Zhang, Chengqi and Philip, S Yu},
  journal=IEEE_J_NNLS,
  volume={32},
  number={1},
  pages={4--24},
  year={2020},

}

@inproceedings{cao2022cross,
  title={Cross vision-{RF} gait re-identification with low-cost {RGB}-d cameras and mm{W}ave radars},
  author={Cao, Dongjiang and Liu, Ruofeng and Li, Hao and Wang, Shuai and Jiang, Wenchao and Lu, Chris Xiaoxuan},
  booktitle={Proc. ACM IMWUT},
  volume={6},
  number={3},
  pages={1--25},
  year={2022},

}

@article{crossharhong2024,
  title={Crosshar: Generalizing cross-dataset human activity recognition via hierarchical self-supervised pretraining},
  author={Hong, Zhiqing and Li, Zelong and Zhong, Shuxin and Lyu, Wenjun and Wang, Haotian and Ding, Yi and He, Tian and Zhang, Desheng},
  journal={Proc. ACM IMWUT},
  volume={8},
  number={2},
  pages={1--26},
  year={2024},
}

@inproceedings{zhang2025mmhiu,
  title={mm{HIU}: a human-to-human interaction understanding system based on mm{W}ave sensing},
  author={Zhang, Fenglin and Wu, Chenglin and Zhou, Anfu and Ma, Huadong},
  booktitle={Proc. IEEE ICASSP},
  pages={1--5},
  year={2025},
}

@inproceedings{alam2021palmar,
  title={{PALMAR}: Towards adaptive multi-inhabitant activity recognition in point-cloud technology},
  author={Alam, Mohammad Arif Ul and Rahman, Md Mahmudur and Widberg, Jared Q},
  booktitle={Proc. IEEE INFOCOM},
  pages={1--10},
  year={2021},

}

@inproceedings{li2021towards,
  title={Towards efficient graph convolutional networks for point cloud handling},
  author={Li, Yawei and Chen, He and Cui, Zhaopeng and Timofte, Radu and Pollefeys, Marc and Chirikjian, Gregory S and Van Gool, Luc},
  booktitle={Proc. IEEE/CVF ICCV},
  pages={3752--3762},
  year={2021}
}

\end{document}